\documentclass[conference]{IEEEtran}

\usepackage{subfig} 
\usepackage{graphicx}
\usepackage{amsmath} 
\usepackage{amssymb} 
\usepackage{changepage}
\usepackage[hyphens]{url}
\usepackage{array}
\newcolumntype{L}{>{\centering\arraybackslash}m{3cm}}

\usepackage{algorithm}
\usepackage{algpseudocode}
\algnewcommand\RETURN{\State \algorithmicreturn}

\usepackage{hyperref}

\begin{document} 

\title{Sampling Method for Fast Training of Support Vector Data Description}

% It is OKAY to include author information, even for blind
% submissions: the style file will automatically remove it for you
% unless you've provided the [accepted] option to the icml2016
% package.

\author{
\IEEEauthorblockN{Arin Chaudhuri and Deovrat Kakde and Maria Jahja and Wei Xiao and \\ Seunghyun Kong and Hansi Jiang
and Sergiy Peredriy}

\IEEEauthorblockA{SAS Institute \\ Cary, NC, USA\\
Email: arin.chaudhuri@sas.com, deovrat.kakde@sas.com, maria.jahja@sas.com, wei.xiao@sas.com\\
seunhyun.kong@sas.com, hansi.jiang@sas.com, sergiy.peredriy@sas.com}
}

\maketitle
\begin{abstract}
Support Vector Data Description (SVDD) is a popular outlier detection technique
which constructs a flexible description of the input data. SVDD computation
time is high for large training datasets which limits its use in big-data
process-monitoring applications. We propose a new iterative sampling-based
method for SVDD training. The method incrementally learns the training data
description at each iteration by computing SVDD on an independent random sample
selected with replacement from the training data set. The experimental results
indicate that the proposed method is extremely fast and provides good data
description.
\end{abstract}
\section{Introduction}

Support Vector Data Description (SVDD) is a machine learning technique used
for single class classification and outlier detection. SVDD technique is
similar to Support Vector Machines and was first introduced by Tax and Duin
\cite{tax2004support}. It can be used to build a flexible boundary around
single class data. Data boundary is characterized by observations designated
as support vectors. SVDD is used in domains where majority of data belongs to
a single class. Several researchers have proposed use of SVDD for multivariate
process control \cite{sukchotrat2009one}. Other applications of SVDD involve
machine condition monitoring \cite{widodo2007support, ypma1999robust} and image
classification \cite{sanchez2007one}.

\subsection{\bf Mathematical Formulation of SVDD}
\label{mfsvdd}
\paragraph*{\bf Normal Data Description}\mbox{}\\
The SVDD model for normal data description builds a minimum radius hypersphere around the data.
\paragraph*{\bf Primal Form}\mbox{}\\
Objective Function:
\begin{equation}
\min R^{2} + C\sum_{i=1}^{n}\xi _{i}, 
\end{equation}
subject to: 
\begin{align}
\|x _{i}-a\|^2 \leq R^{2} + \xi_{i}, \forall i=1,\dots,n,\\
\xi _{i}\geq 0, \forall i=1,...n.
\end{align}
where:\\
$x_{i} \in {\mathbb{R}}^{m}, i=1,\dots,n  $ represents the training data,\\
$ R:$ radius, represents the decision variable,\\
$\xi_{i}:$ is the slack for each variable,\\
$a$: is the center, a decision variable, \\
$C=\frac{1}{nf}:$ is the penalty constant that controls the trade-off between the volume and the errors, and,\\
$f:$ is the expected outlier fraction.
\paragraph*{\bf Dual Form}\mbox{}\\
The dual formulation is obtained using the Lagrange multipliers.\\ 
Objective Function:
\begin{equation} 
\max\ \sum_{i=1}^{n}\alpha _{i}(x_{i}.x_{i}) - \sum_{i,j}^{ }\alpha _{i}\alpha _{j}(x_{i}.x_{j}) ,
\end{equation}
subject to:
\begin{align}
& &  \sum_{i=1}^{n}\alpha _{i}  = 1,\label{sv:s}\\
& & 0 \leq  \alpha_{i}\leq C,\forall i=1,\dots,n.
\end{align}
where:\\
$\alpha_{i}\in \mathbb{R}$: are the Lagrange constants,\\
$C=\frac{1}{nf}:$ is the penalty constant.
\paragraph*{\bf Duality Information}\mbox{}\\
Depending upon the position of the observation, the following results hold:
Center Position: \begin{equation} \sum_{i=1}^{n}\alpha _{i}x_{i}=a. \label{sv:0} \end{equation}
Inside Position: \begin{equation} \left \| x_{i}-a \right \| < R \rightarrow \alpha _{i}=0.\end{equation}
Boundary Position: \begin{equation} \left \| x_{i}-a \right \| = R \rightarrow 0< \alpha _{i}< C.\end{equation}
Outside Position: \begin{equation}\left \| x_{i}-a \right \| > R \rightarrow \alpha _{i}= C. \label{sv:1} \end{equation}
The radius of the hypersphere is calculated as follows:\\
\begin{equation}   
R^{2}=(x_{k}.x_{k})-2\sum_{i}^{ }\alpha _{i}(x_{i}.x_{k})+\sum_{i,j}^{ }\alpha _{i}\alpha _{j}(x_{i}.x_{j}).
\label{eq:a}
\end{equation}
 using any $ x_{k} \in SV_{<C} $
, where $SV_{<C}$  is the set of support vectors that have $ \alpha _{k} < C $.

\paragraph*{\bf Scoring}\mbox{}\\
For each observation $ z $  in the scoring data set, the distance $ \textrm{dist}^{2}(z) $ is calculated as: 
\begin{equation}  dist^{2}(z)=(z.z) - 2\sum_{i}^{ }\alpha _{i}(x_{i}.z) +\sum_{i,j}^{ }\alpha _{i}\alpha _{j}(x_{i}.x_{j}) \end{equation}
and observations with $ dist^{2}(z) > R^{2} $ are designated as outliers.

The spherical data boundary can include a significant amount of space with a very sparse distribution of training
observations which leads to a large number of falses positives. The use of kernel functions leads to better compact
representation of the training data.
\paragraph*{\bf Flexible Data Description}\mbox{}\\
The Support Vector Data Description is made flexible by replacing the inner product $ (x_{i}.x_{j}) $ in equation \eqref{eq:a} with a 
suitable kernel function $ K(x_{i},x_{j}) $. The Gaussian kernel function used in this paper is defined as:
\begin{equation}  
K(x_{i}, x_{j})= \exp  \dfrac{ -\|x_i - x_j\|^2}{2s^2}
\label{eq:b}
\end{equation}
where $s$: Gaussian bandwidth parameter.

The modified mathematical formulation of SVDD with kernel function is:

Objective function:
\begin{equation}  \label{eq:1}
\max\ \sum_{i=1}^{n}\alpha _{i}K(x_{i},x_{i}) - \sum_{i,j}^{ }\alpha _{i}\alpha _{j}K(x_{i},x_{j}),
\end{equation}
Subject to:
\begin{align}
& &\sum_{i=1}^{n}\alpha _{i} = 1, \label{eq:2} \\
& & 0 \leq  \alpha_{i}\leq C = \frac{1}{nf} , \forall i=1,\dots,n. \label{eq:3}
\end{align}
Conditions similar to \eqref{sv:0} to \eqref{sv:1} continue to hold even when the kernel function is used.\\
The threshold $R^{2}$ is calculated as :
\begin{multline}
R^{2} = K(x_{k},x_{k})-2\sum_{i}^{ }\alpha _{i}K(x_{i},x_{k})+\sum_{i,j}^{ }\alpha _{i}\alpha _{j}K(x_{i},x_{j})
\end{multline}
using any $ x_{k} \in SV_{<C} $
, where $SV_{<C}$  is the set of support vectors that have $ \alpha _{k} < C $.

\paragraph*{\bf Scoring}\mbox{}\\
For each observation $z$ in the scoring dataset, the distance $ dist^{2}(z) $ is calculated as: 
\begin{equation} dist^{2}(z)= K(z,z) - 2\sum_{i}^{ }\alpha _{i}K(x_{i},z) +\sum_{i,j}^{ }\alpha _{i}\alpha _{j}K(x_{i},x_{j}),\end{equation}
and the observations with $ dist^{2}(z) > R^{2} $ are designated as outliers.

\section{Need for a Sampling-based Approach}
\begin{figure}[h]
	\centering
	\includegraphics[scale=0.25]{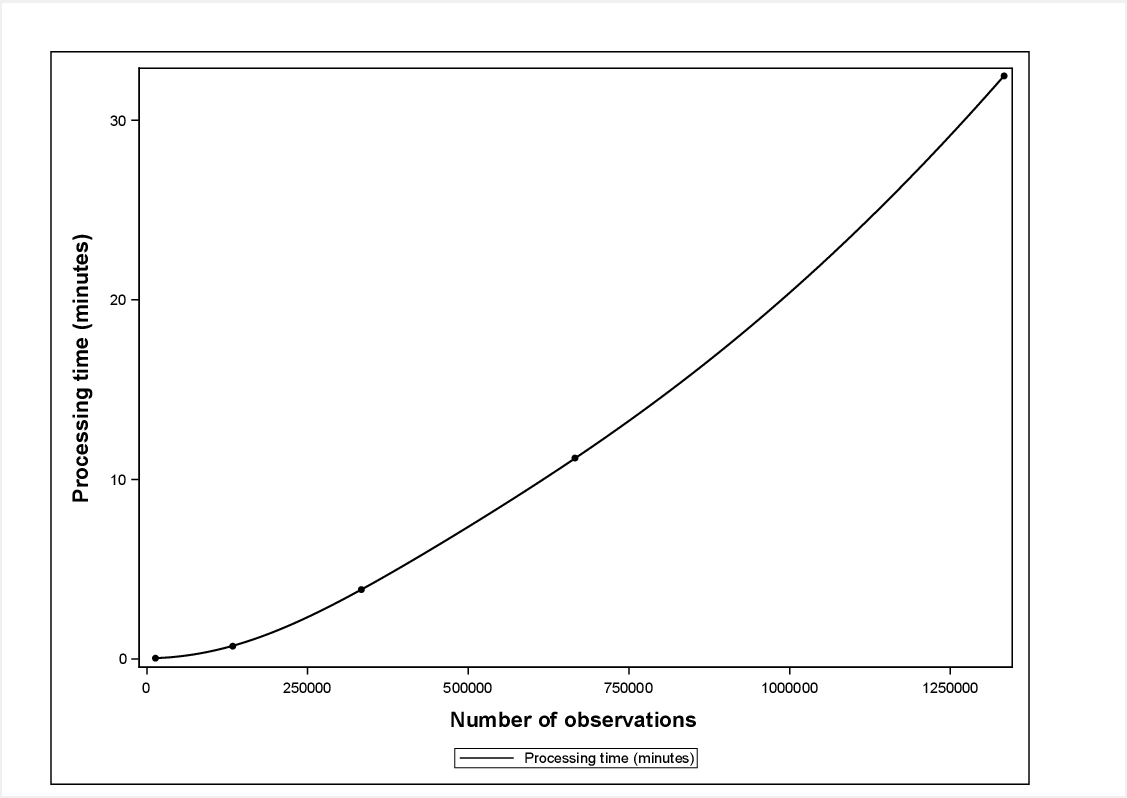}
	\caption{SVDD Training Time: Two Donut data}
	\label{fig:image_0}
\end{figure}

As outlined in Section \ref{mfsvdd}, SVDD of a data set is obtained
by solving a quadratic programming problem. The time required to solve
the quadratic programming problem is directly related to the number of
observations in the training data set. The actual time complexity depends
upon the implementation of the underlying Quadratic Programming solver.
We used LIBSVM  to evaluate SVDD training time as
a function of the training data set size.
We have used C++ code that uses LIBSVM ~\cite{chang2011libsvm} implementation of SVDD
the examples in this paper, we have also provided a Python implmentation which uses Scikit-learn~\cite{scikit-learn} 
at \cite{smsvddg}.
Figure~\ref{fig:image_0} shows
processing time as a function of training data set size for the two donut
data set (see Figure \ref{fig:image_3} for a scatterplot of the two donut
data). In Figure~\ref{fig:image_0} the x-axis indicates the training data set
size and the y-axis indicates processing time in minutes. As indicated in
Figure~\ref{fig:image_0}, the SVDD training time is low for small or moderately
sized training data but gets prohibitively high for large datasets.

There are applications of SVDD in areas such as process control and equipment
health monitoring where size of training data set can be very large, consisting
of few million observations. The training data set consists of sensors readings
measuring multiple key health or process parameters at a very high frequency.
For example, a typical airplane currently has $\approx$7,000 sensors
measuring critical health parameters and creates 2.5 terabytes of data per day.
By 2020, this number is expected to triple or quadruple to over 7.5 terabytes
\cite{ege2015IoT}. In such applications, multiple SVDD training models are
developed, each representing separate operating mode of the equipment or process
settings. The success of SVDD in these applications require algorithms which can
train using huge amounts of training data in an efficient manner.

To improve performance of SVDD training on large data sets, we propose a new
sampling based method. Instead of using all observations from the training data
set, the algorithm computes the training data SVDD by iteratively computing SVDD
on independent random samples obtained from the training data set and combining
them. The method works well even when the random samples have few observations.
We also provide a criteria for detecting convergence. At convergence the our
method provides a data description that compares favorably with result obtained
by using all the training data set observations.

The rest of this document is organized as follows: Section~\ref{sbm} provides
details of the proposed sampling-based iterative method. Results of training
with the proposed method are provided in section~\ref{res}; the analysis of high
dimensional data is provided in section~\ref{pd}; the results of a simulation
study on random polygons is provided in section Section~\ref{ss} and we provide
our conclusions in section~\ref{cn}.

\textbf{Note:\underline{\underline{}}} \textit{In the remainder of this paper, we refer to the training method using all
observations in one iteration as the full SVDD method.\textit{}}

\section{Sampling-based Method}
\label{sbm}
The Decomposition and Combination method of Luo et.al.\cite{luo2010fast} and
K-means Clustering Method of Kim et.al.\cite{kim2007fast}, both use sampling for
fast SVDD training, but are computationally expensive. The first method by Lou
et.al. uses an iterative approach and requires one scoring action on the entire
training data set per iteration. The second method by Kim et.al. is a classic
divide and conquer algorithm. It uses each observation from the training data
set to arrive at the final solution.
%Additional details about these methods are provided in Appendix \ref{apd}.

In this section we describe our sampling-based method for fast SVDD training.
The method iteratively samples from the training data set with the objective
of updating a set of support vectors called as the master set of support
vectors ($SV^{*}$). During each iteration, the method updates $SV^{*}$ and
corresponding threshold $R^{2}$ value and center $a$. As the threshold value
$R^{2}$ increases, the volume enclosed by the $SV^{*}$ increases. The method
stops iterating and provides a solution when the threshold value $R^{2}$ and the
center $a$ converge. At convergence,
the members of the master set of support vectors $SV^{*}$, characterize
the description of the training data set. For all test cases, our method
provided a good approximation to the solution that can be obtained by using all
observations in the training data set.

Our method addresses drawbacks of existing sampling based methods proposed
by Luo et.al.\cite{luo2010fast} and Kim et.al.\cite{kim2007fast}. In each
iteration, our method learns using very a small sample from the training data
set during each step and typically uses a very small subset of the training data set. The method
does not require any scoring actions while it trains.

The sampling method works well for different sample sizes for the random draws in the iterations. 
It also provides a better alternative to training SVDD on one large random sample from the training data
set, since establishing a right size, especially with high dimensional data, is
a challenge.

The important steps in this algorithm are outlined below:\mbox{}\\
\textbf{Step 1:}
The algorithm is initialized by selecting a random sample $S_{0}$ of size $n$
from the training data set of $M$ observations ($n \ll M$). SVDD of $S_{0}$ is
computed to obtain the corresponding set of support vectors $SV_{0}$. The set
$SV_{0}$ initializes the master set of support vectors $SV^{*}$. The iteration
number $i$ is set to 1.\\
\textbf{Step 2:}
During this step, the algorithm updates the master set of support vectors,
$SV^{*}$ until the convergence criteria is satisfied. In each iteration $i$,
following steps are executed:
\begin{adjustwidth}{2mm}{0pt} \textbf{Step 2.1:}  A random sample $S_{i}$ of size $n$ is selected and its SVDD is computed. The corresponding support vectors are designated as $SV_{i}$.\\
\textbf{Step 2.2}: A union of $SV_{i}$ with the current master set of support vectors,  $SV^{*}$ is taken to obtain a set $S_{i}^{'}$ ($S_{i}^{'}=SV_{i} \bigcup SV^{*}$).\\
\textbf{Step 2.3: }SVDD of $S_{i}^{'}$ is computed to obtain corresponding support vectors $SV_{i}^{'}$, threshold value
$R_{i}^{2}$ and ``center'' $a_{i}$ (which we define as $\sum_{i}\alpha_i x_i$ even when a Kernel is used).  The set $SV_{i}^{'}$, is designated as the new master set of support vectors $SV^{*}$.  
\end{adjustwidth}
\textbf{ Convergence Criteria:} 
At the end of each iteration $i$, the following conditions are checked to determine convergence.
\begin{adjustwidth}{2mm}{0pt}
\begin{enumerate}
\item  $i$ = $maxiter$, where $maxiter$ is the maximum number of iteration; or\\
\item $ \| a_{i} - a_{i-1} \| \le  \epsilon_1 \|a_{i-1}\|$,  and
   $\left \| R_{i}^{2}-R_{i-1}^{2} \right \| \le \epsilon_2 R_{i-1}^{2}$ where $\epsilon_1,\epsilon_2$ are appropriately
chosen tolerance parameters.
\end{enumerate}
\end{adjustwidth}
If the maximum number of iterations is reached or the second condition satisfied for $t$ consecutive iterations,
convergence is declared. In many cases checking the convergence of just $R_i^2$ suffices.

The pseudo-code for this method is provided in algorithm~\ref{alg:the_alg1}. The pseudo-code uses following notations:
\begin{enumerate}
\item $S_{i} \leftarrow SAMPLE (T, n)$ denotes the data set $S_{i}$ obtained by selecting random sample of size $n$ from data set $T$.
\item $\delta S_{i}$ denotes SVDD computation on data set $S_{i}$.
\item $\langle SV_{i}, R_{i}^{2}, a_{i} \rangle \leftarrow \delta S_{i} $ denotes the set of support vectors $SV_{i}$, threshold value $R_{i}^{2}$ and center $a_{i}$ obtained by performing SVDD computations on data set $S_{i}$.
\end{enumerate}
\begin{algorithm}
	\caption{Sampling-based iterative method}\label{euclid}
	\label{alg:the_alg1}
	\begin{algorithmic}[1]
		\State  $T$ (training data set) , $n$ (sample size), convergence criteria, $s$ (Gaussian bandwidth parameter),
$f$ (fraction of outliers) and $t$ (required number of consecutive observations satisfying convergence criteria ).
		\State  $S_{0} \leftarrow SAMPLE(T, n)$
		\State  $ \langle SV_{0}, R_{0}^{2}, a_{0} \rangle \leftarrow \delta S_{0}$ 
		\State  $SV^{*} \leftarrow SV_{0}$
		\State  $i=1$
		\While {(Convergence criteria not satisfied for $t$ consecutive obs)} 
		\State  $S_{i} \leftarrow SAMPLE(T, n)$
		\State  $\langle SV_{i}, R_{i}^{2}, a_{i} \rangle \leftarrow \delta S_{i}$
		\State $S_{i}^{'} \leftarrow SV_{i} \bigcup SV^{*}$.		
		\State  $\langle SV_{i}^{'}, R_{i}^{2'}, a_{i}^{'}\rangle \leftarrow \delta S_{i}^{'}$
		\State Test for convergence
		\State $SV^{*} \leftarrow SV_{i}^{'}$
		\State $i=i+1$
		\EndWhile
		\RETURN { $SV^{*}$}
		
	\end{algorithmic}
\end{algorithm}

As outlined in steps 1 and 2, the algorithm obtains the final training data
description by incrementally updating the master set of support vectors $SV^{*}$.
During each iteration, the algorithm first selects a small random
sample $S_{i}$, computes its SVDD and obtains corresponding set of support
vectors, $SV_{i}$. The support vectors of set $SV_{i}$ are included in the
master set of support vectors $SV^{*}$ to obtain $S_{i}^{'}$ ($S_{i}^{'}=SV_{i}
\bigcup SV^{*}$). The set $S_{i}^{'}$ thus represents an incremental expansion
of the current master set of support vectors $SV^{*}$. Some members of $SV_{i}$
can be potentially ``inside'' the data boundary characterized by $SV^{*}$ the
next SVDD computation on $S_{i}^{'}$ eliminates such ``inside'' points.
During initial iterations as $SV^{*}$ gets updated, its threshold value $R_{i}^{2'}$ typically increases and 
the master set of support vectors expands to describe the entire data set.

Each iteration of our algorithm involves two small SVDD computations and one union operation.  The first SVDD
computation is fast since it is perfomed on a small sample of training data set. For the remaining two operations, our
method exploits the fact that for most data sets support vectors obtained from SVDD are a tiny fraction
of the input data set and both the union operation and the second SVDD computation are fast. So our method consists
of three fast operations per iteration. For most large datasets we have experimented on the time to convergence is fast
and we achieve a reasonable approximation to full SVDD in a fraction to time needed compute SVDD with the full dataset.

\subsubsection{Distributed Implementation}
\begin{figure}[h]
	\centering
	\includegraphics[scale=0.25]{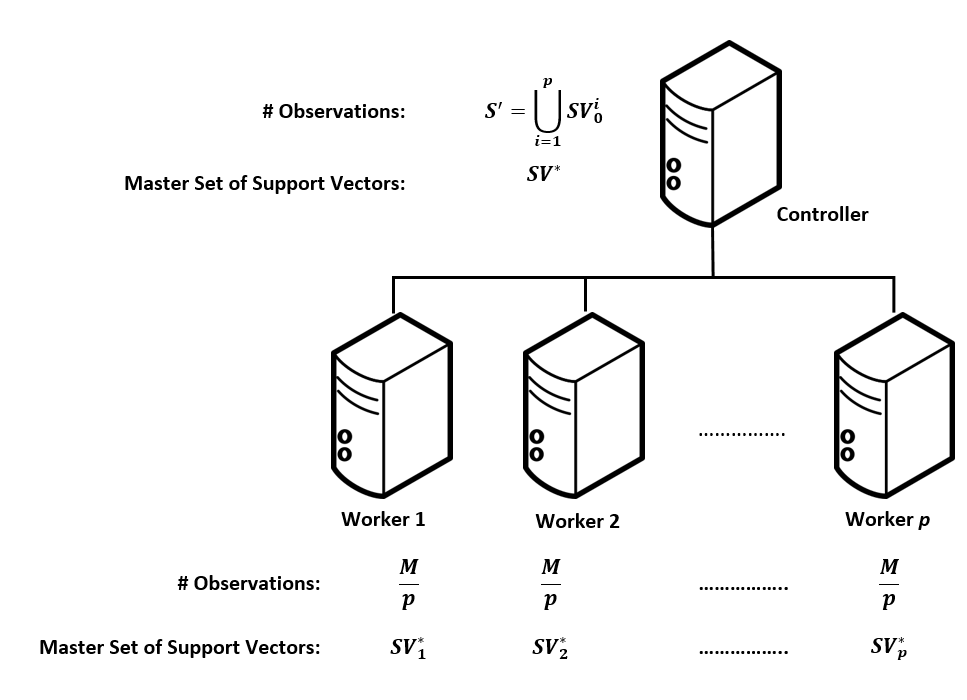}
	\caption{Distributed Implementation}
	\label{fig:image_511}
\end{figure}

For extremely large training datasets, efficiency gains using distributed
implementation are possible. Figure \ref{fig:image_511} describes SVDD solution
using the sampling method outlined in section \ref{sbm} utilizing a distributed
architecture. The training data set with $M$ observations is first distributed
over $p$ worker nodes. Each worker node computes SVDD of its $\dfrac{M}{p}$
observations using the sampling method to obtain its own master set of support
vectors $SV_{i}^*$. Once SVDD computations are completed, each worker node
promotes its own master set of support vectors $SV_{i}^*$, to the controller
node. The controller node takes a union of all worker node master sets of
support vectors, $SV_{i}^*$ to create data set $S^{'}$. Finally, solution is
obtained by performing SVDD computation on $S^{'}$. The corresponding set of
support vectors $SV^{*}$ are used to approximate the original training data set
description.

\section{Results}
\label{res}
\begin{figure}
	\centering     %%% not \center
	\subfloat[Banana-shaped data]{\label{fig:image_1}\includegraphics[scale=0.11]{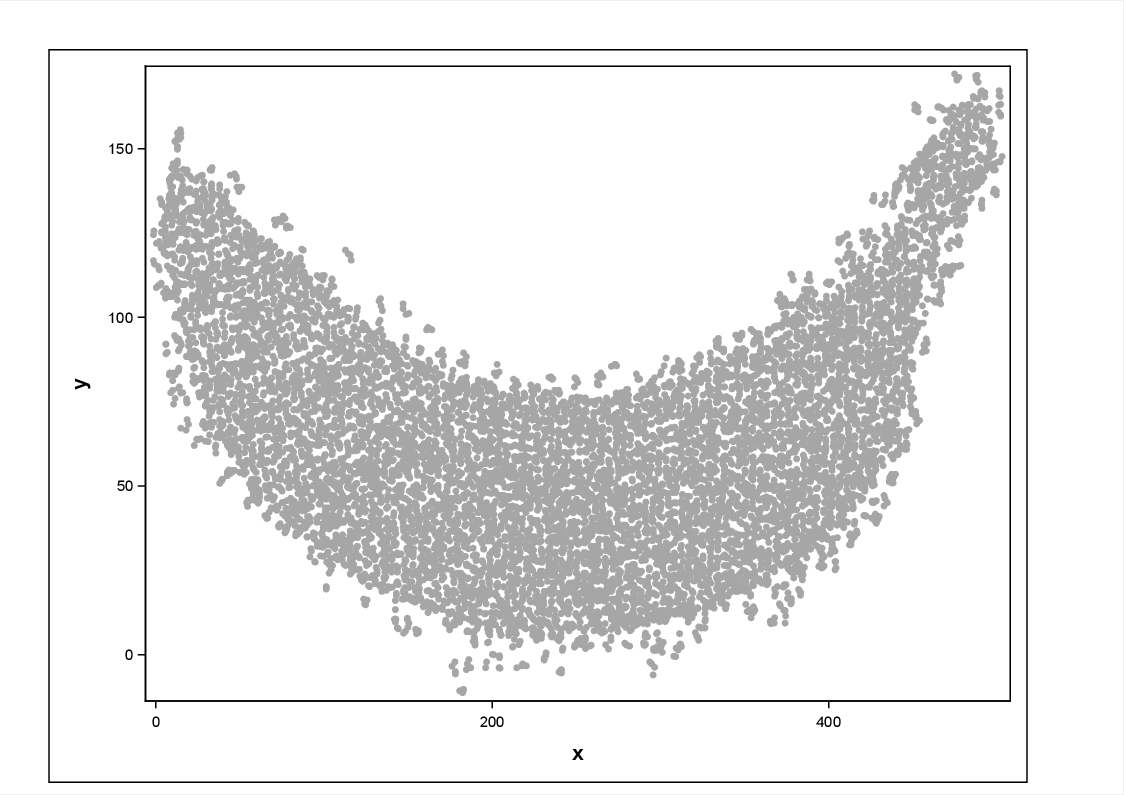}}
	\subfloat[Star-shaped data]{\label{fig:image_2}\includegraphics[scale=0.11]{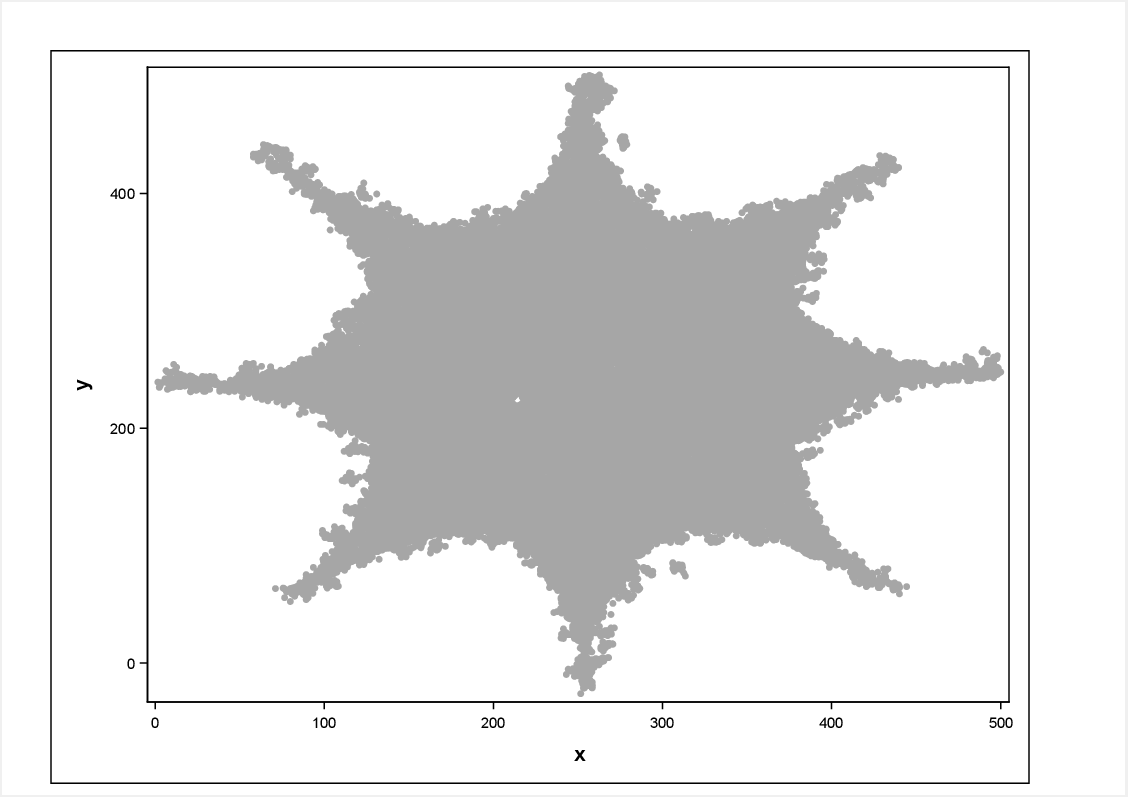}}
	\subfloat[Two-donut-shaped data]{\label{fig:image_3}\includegraphics[scale=0.11]{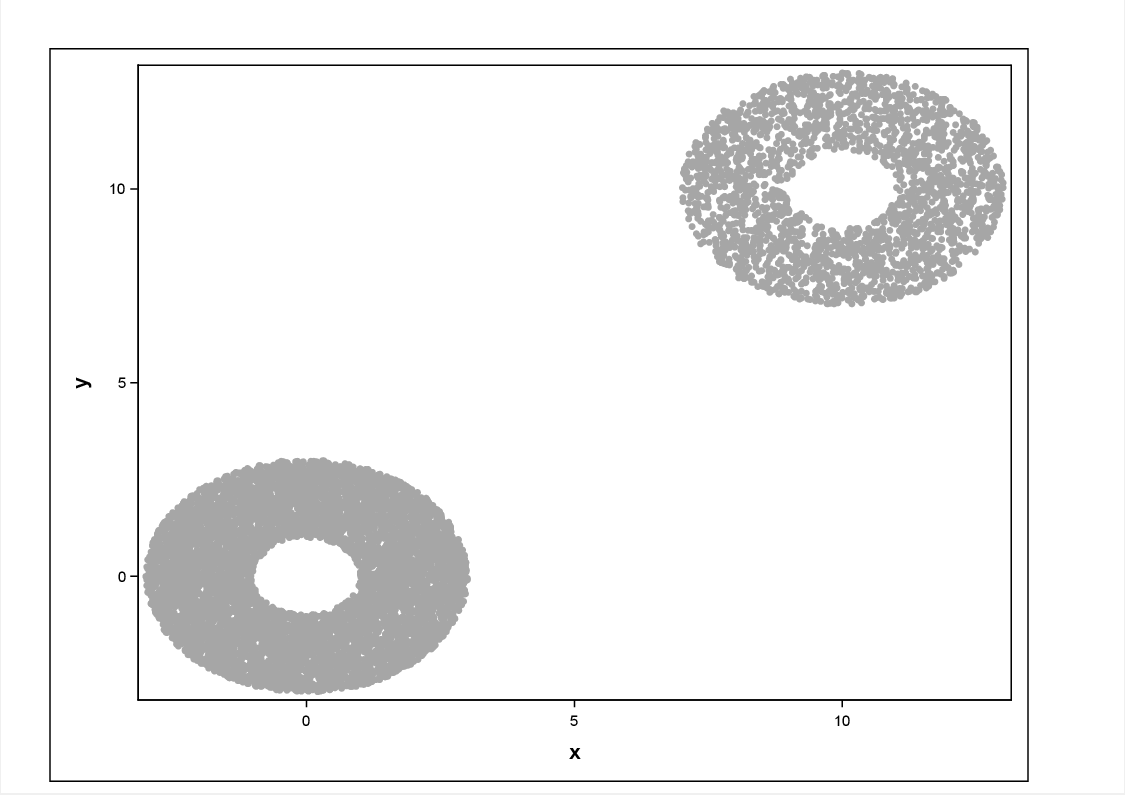}}
	\caption{Scatter plots}\label{fig:image_4}
\end{figure}

\begin{figure}
\centering     %%% not \center
\subfloat[Run time vs. sample size]{\label{fig:image_51}\includegraphics[width=3.00in]{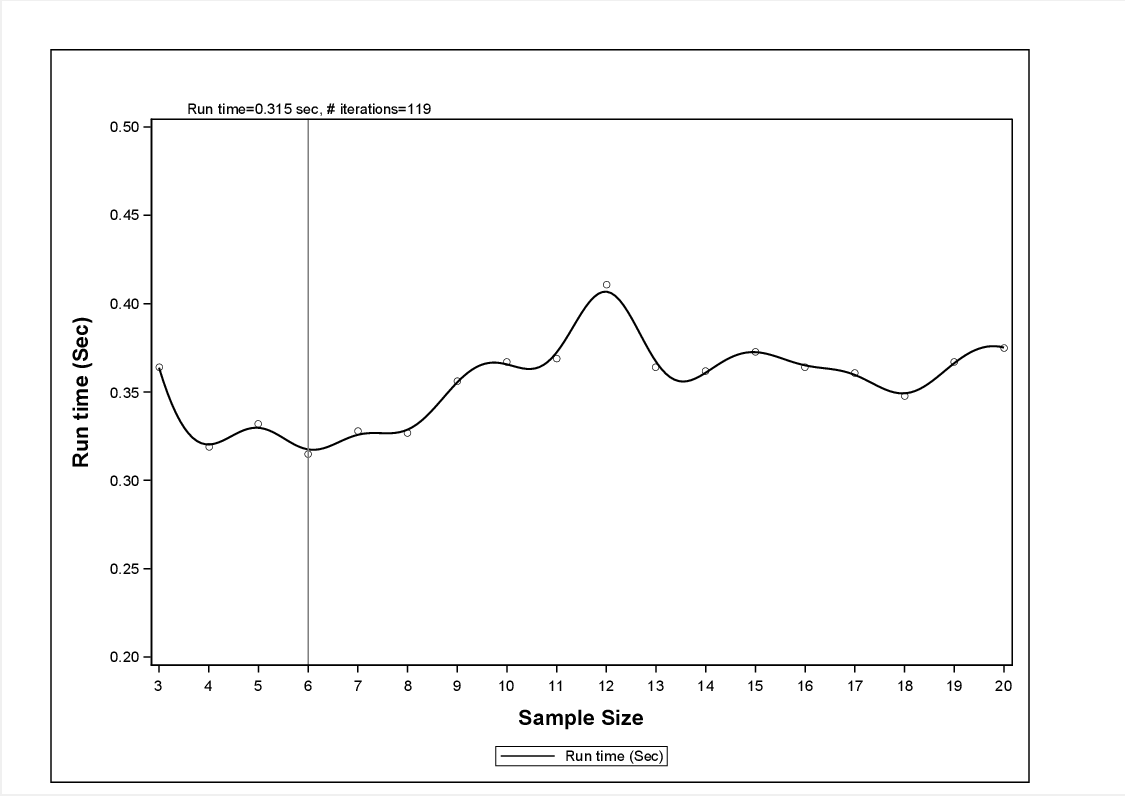}}\mbox{}\\
\subfloat[\# iterations vs. sample size]{\label{fig:image_52}\includegraphics[width=3.00in]{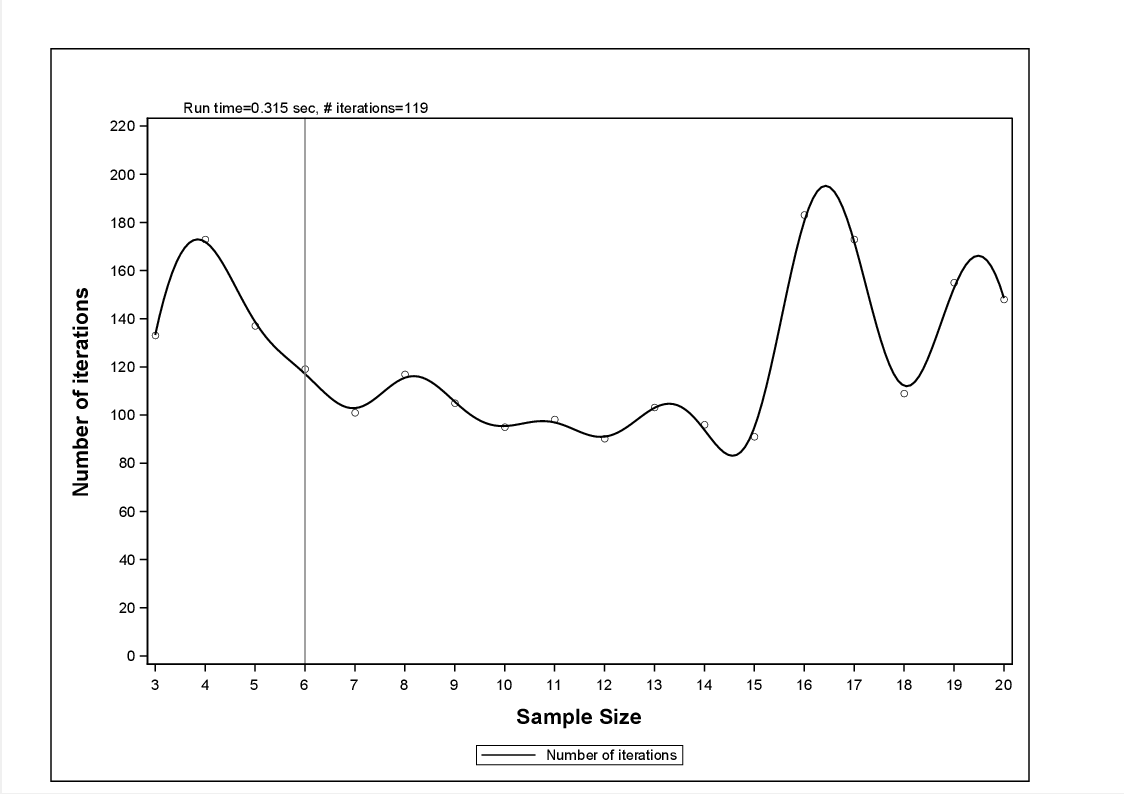}}
\caption{Banana-shaped data}\label{fig:image_5}
\end{figure}

\begin{figure}
\centering     %%% not \center
\subfloat[Run time vs. sample size]{\label{fig:image_61}\includegraphics[width=3.00in]{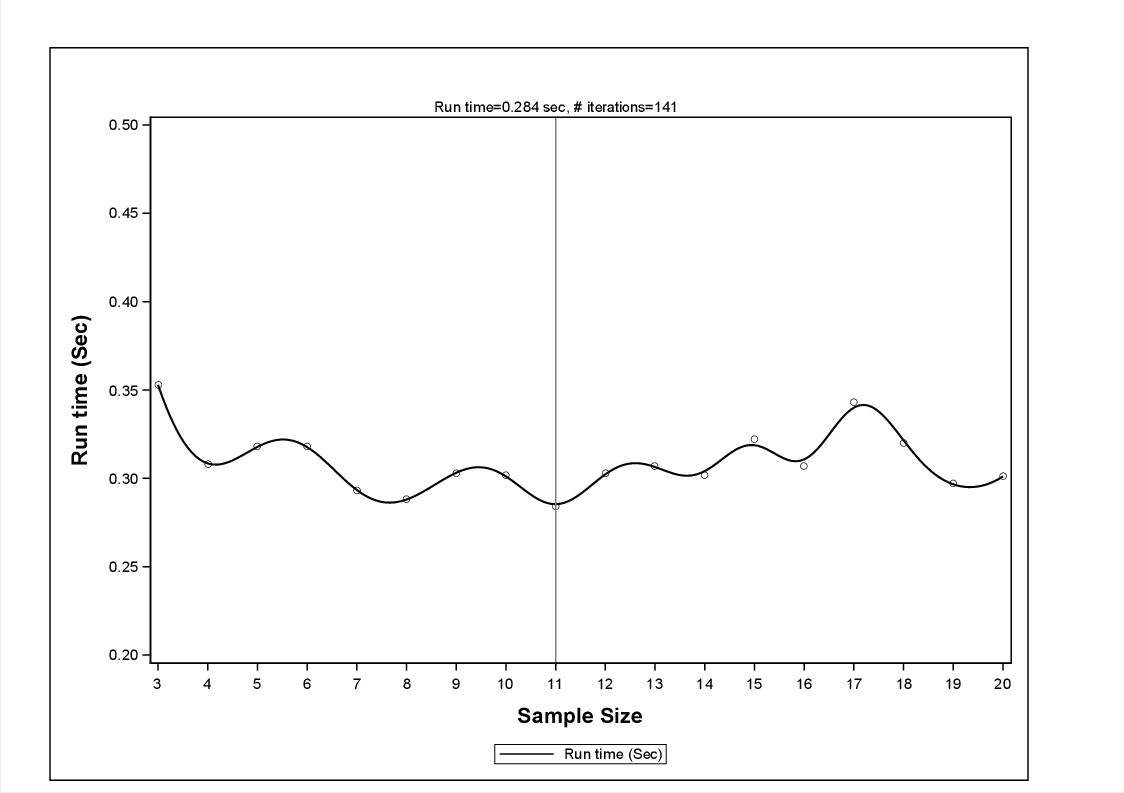}}\mbox{}\\
\subfloat[\# iterations vs. sample size]{\label{fig:image_62}\includegraphics[width=3.00in]{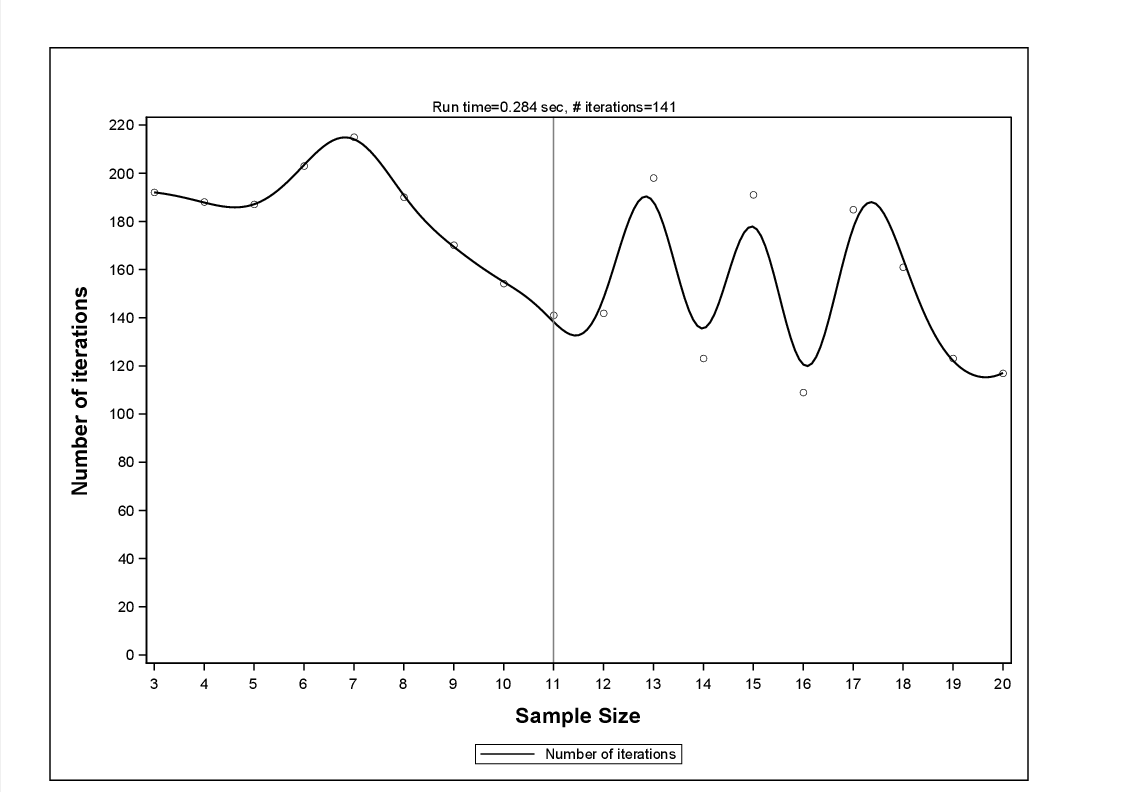}}
\caption{Star-shaped data}\label{fig:image_6}
\end{figure}

\begin{figure}
\centering     %%% not \center
\subfloat[Run time vs. sample size]{\label{fig:image_71}\includegraphics[width=3.00in]{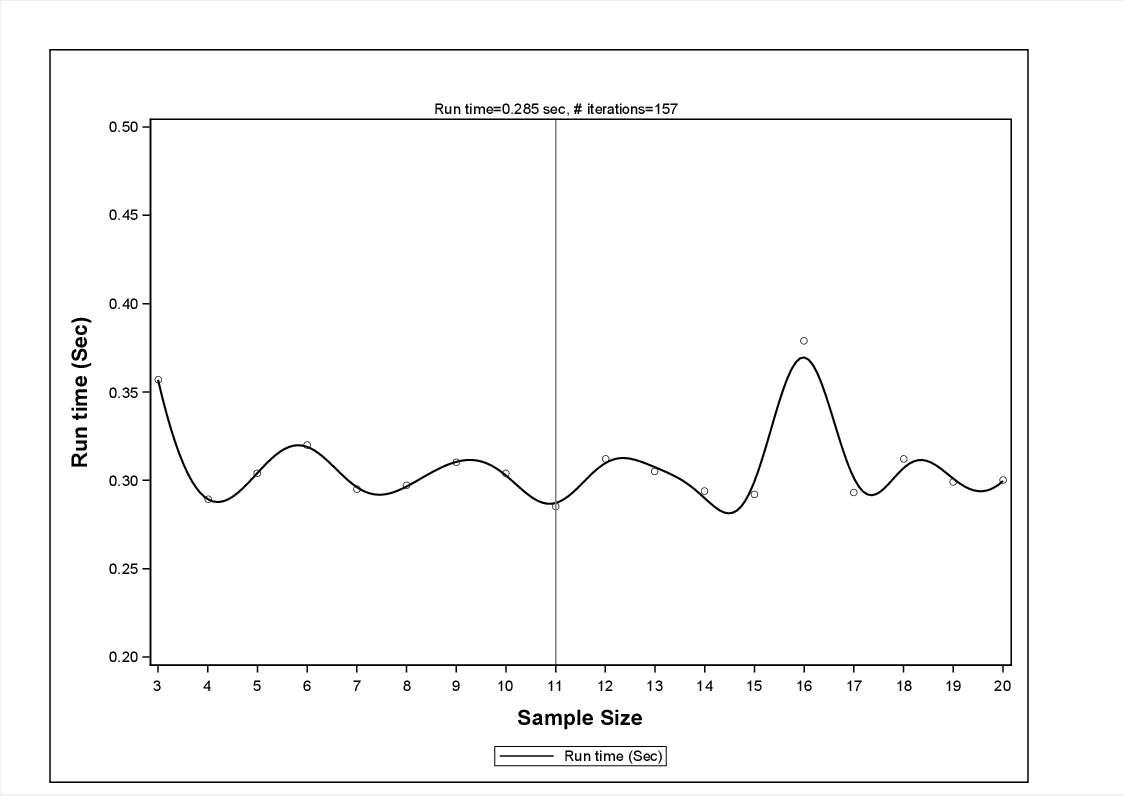}}\\
\subfloat[\# iterations vs. sample size]{\label{fig:image_72}\includegraphics[width=3.00in]{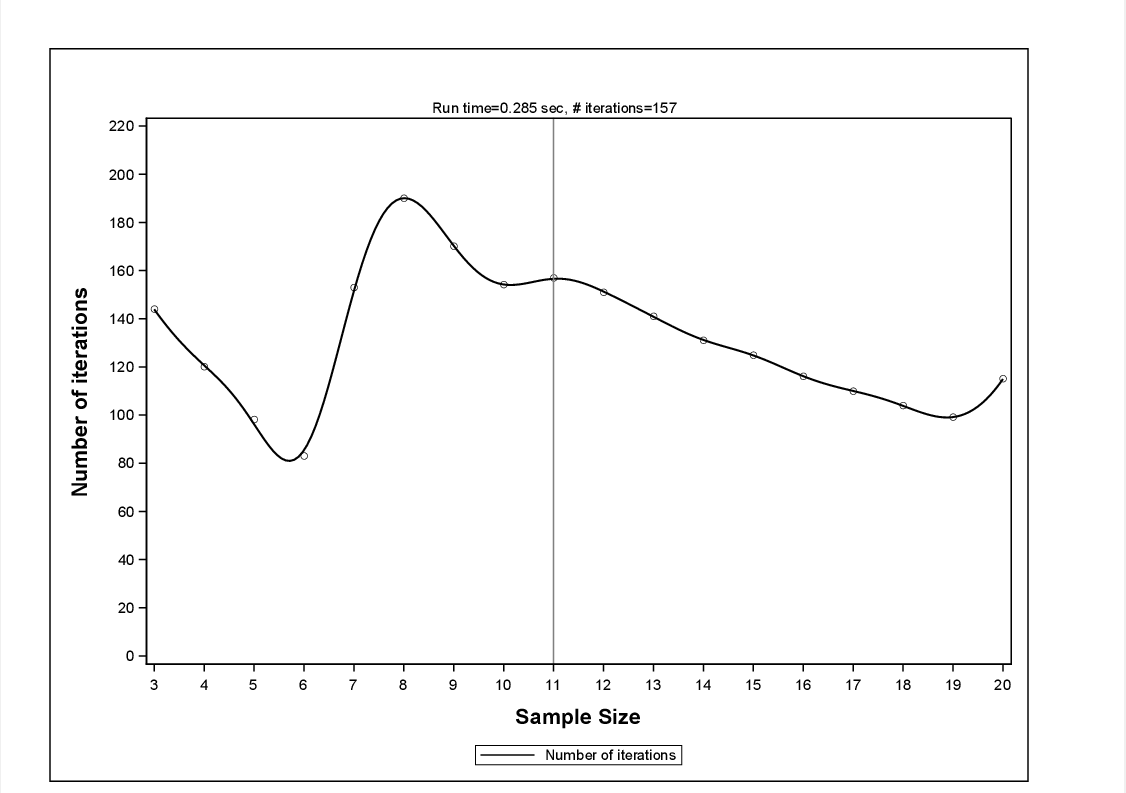}}
\caption{Two Donut data}\label{fig:image_7}
\end{figure}

\begin{figure}[h]
\centering
\includegraphics[width=3.00in]{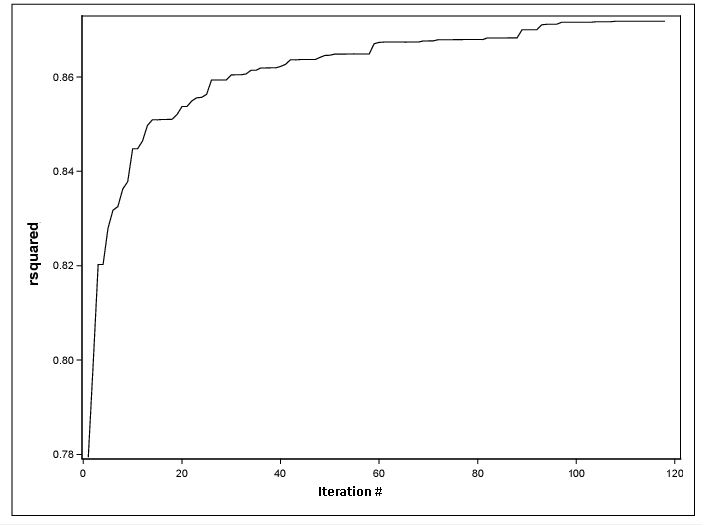}
\caption{Plot of threshold $R^2$ - Banana shaped data (Sample size = 6) }
\label{fig:image_106}
\end{figure}

To test our method we experimented with three data sets of known geometry which we call the Banana-shaped, Star-shaped,
and Two-Donut-shaped
data. The figures \ref{fig:image_1}-\ref{fig:image_3} illustrate these three data sets. 
For each data set, we first obtained SVDD using all observations.
Table~\ref{table:t2} summarizes the results.\\
For each data set, we varied the value of the sample size $n$ from 3 to 20
and obtained multiple SVDD using the sampling method. For each sample size
value, the total processing time and number of iterations till convergence was
noted. Figures \ref{fig:image_5} to \ref{fig:image_7} illustrate the results.
The vertical reference line indicates the sample size corresponding to the
minimum processing time. Table~\ref{table:t3} provides the minimum processing
time, corresponding sample size and other details for all three data sets.
Figure \ref{fig:image_106} shows the convergence of threshold $R^2$ for the
Banana-shaped data trained using sampling method.

\begin{figure}
\begin{tabular}{cc}
\includegraphics[width=1.5in]{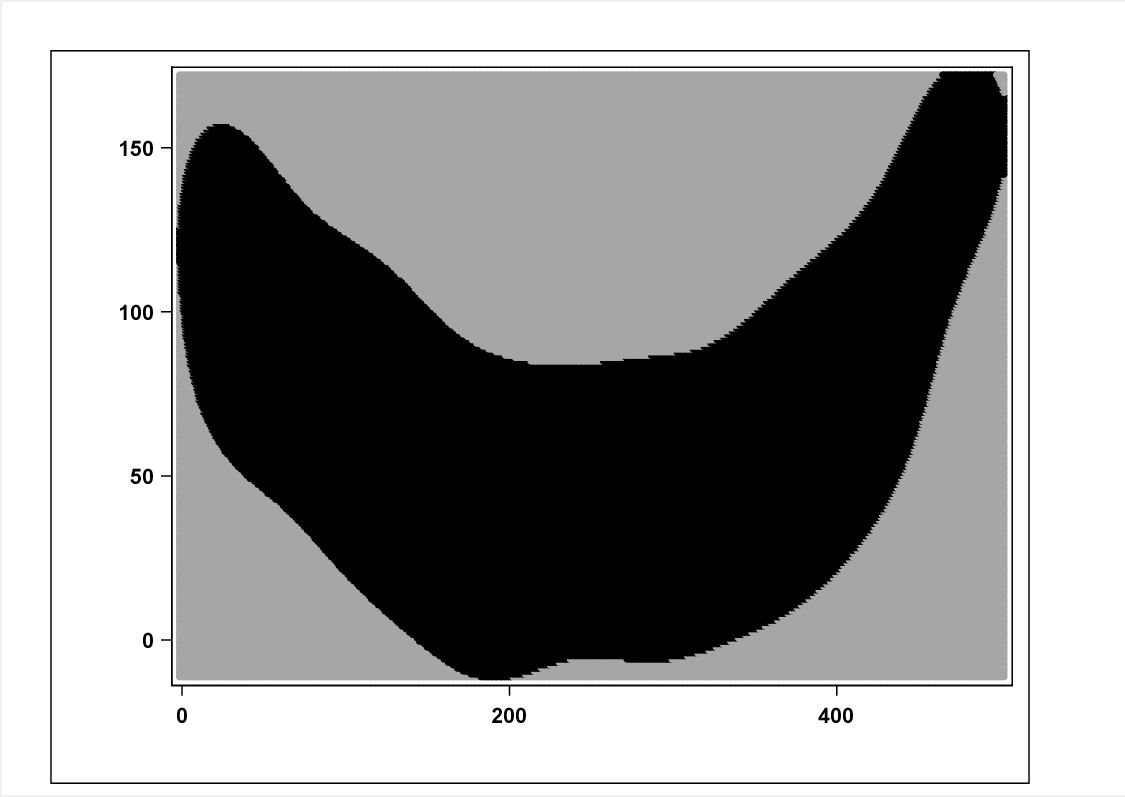}  &   \includegraphics[width=1.5in]{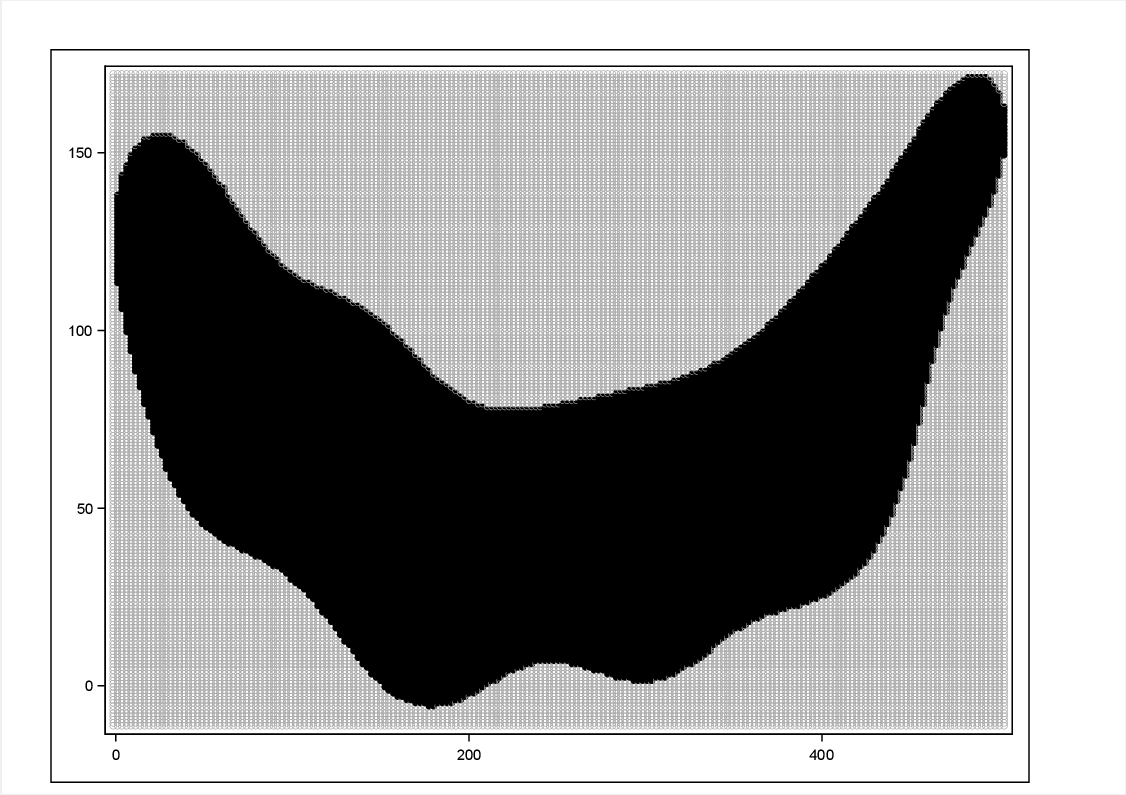} \\
(a)  & (b)  \\[6pt]
\includegraphics[width=1.5in]{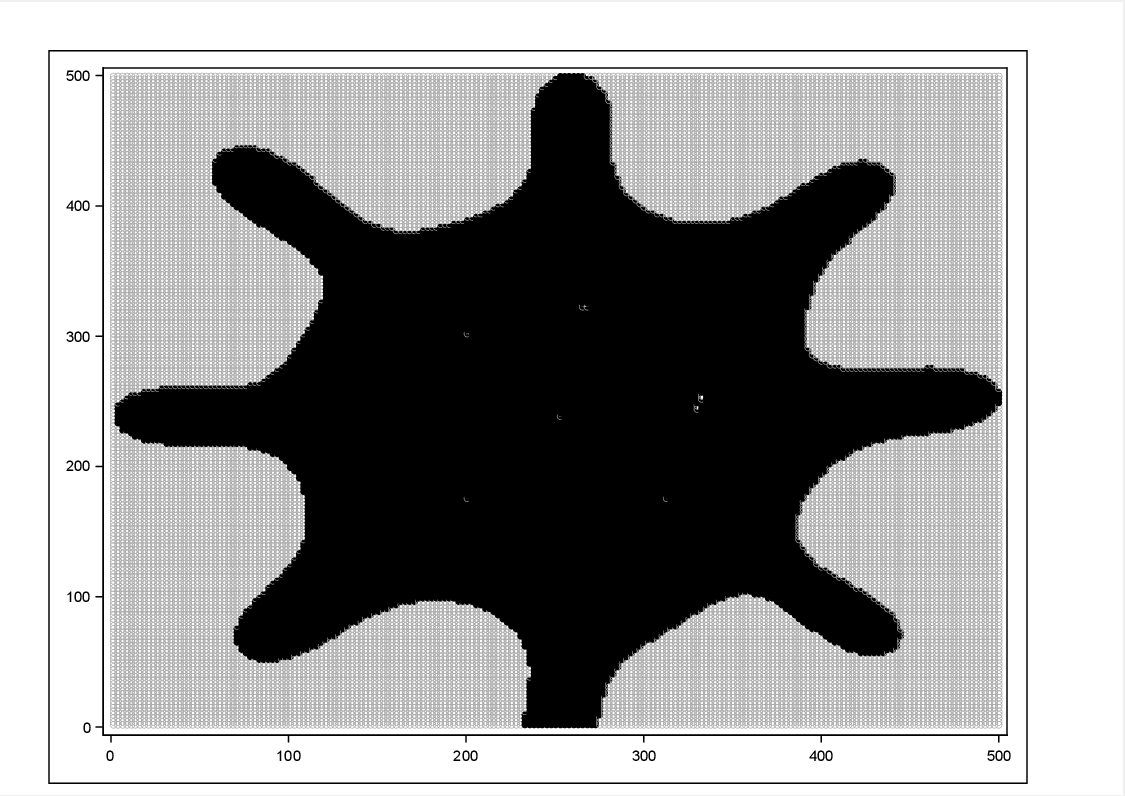} &   \includegraphics[width=1.5in]{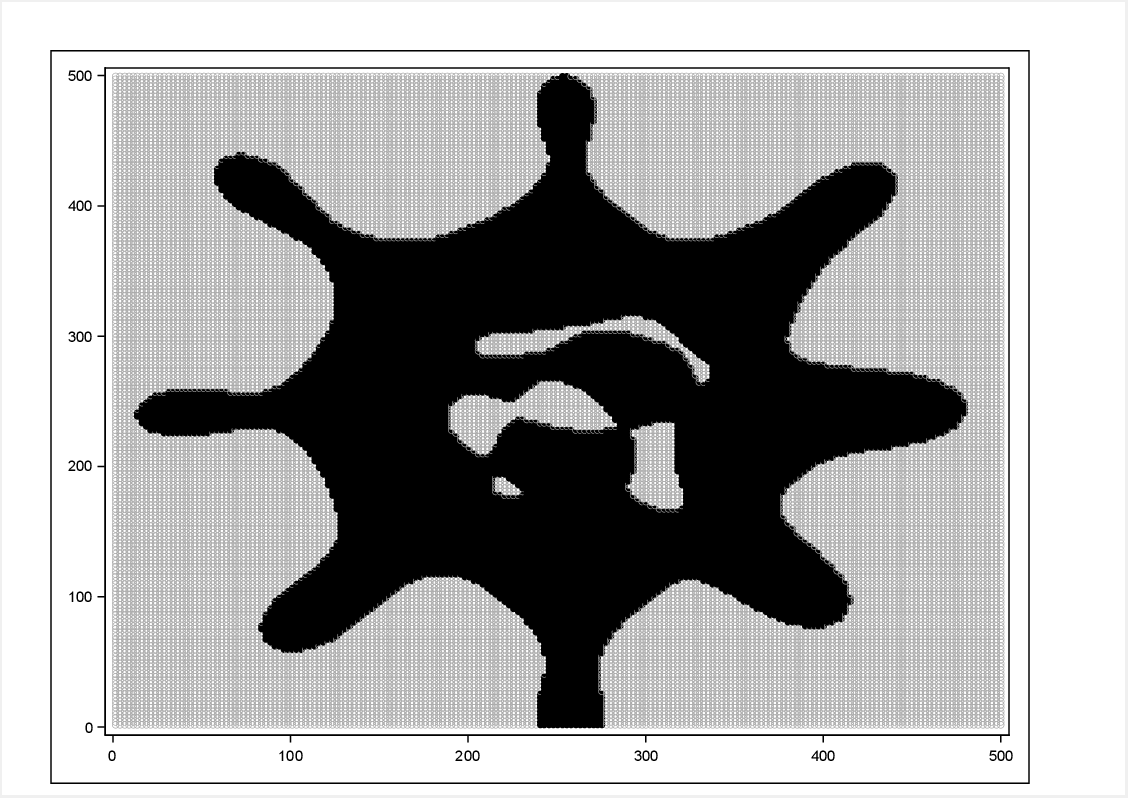} \\
(a)  & (b)  \\[6pt]
\includegraphics[width=1.5in]{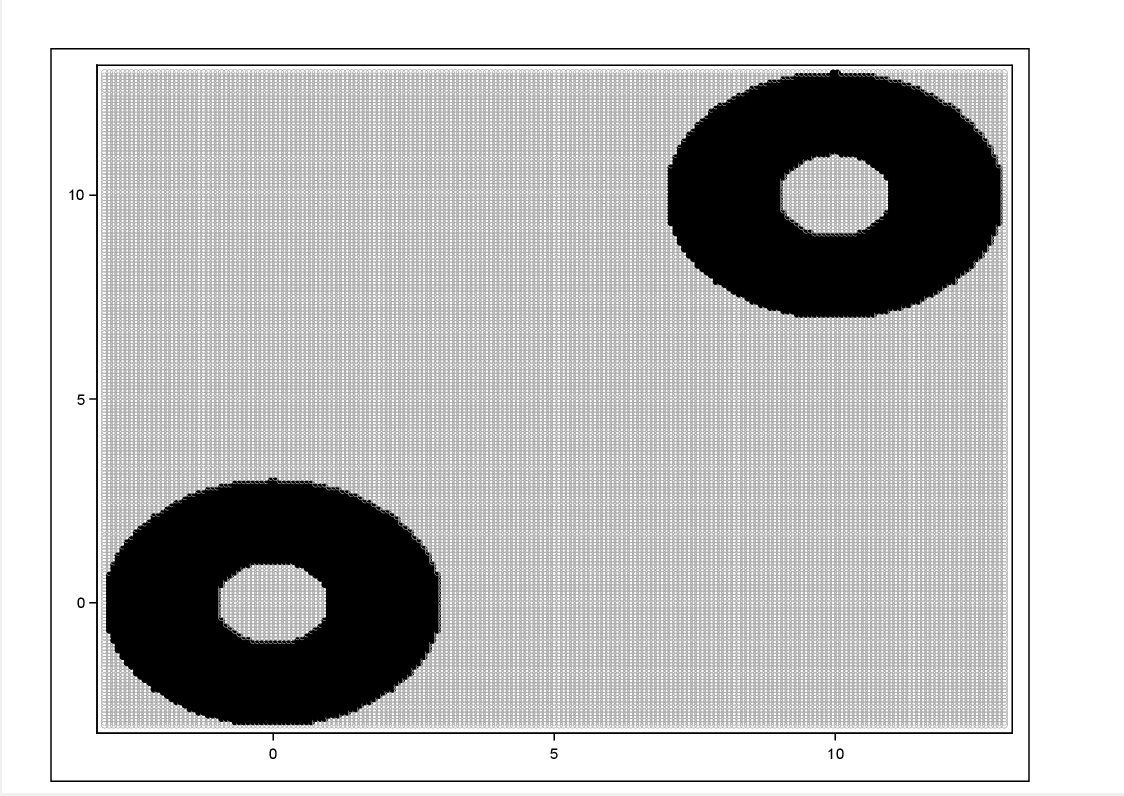} &   \includegraphics[width=1.5in]{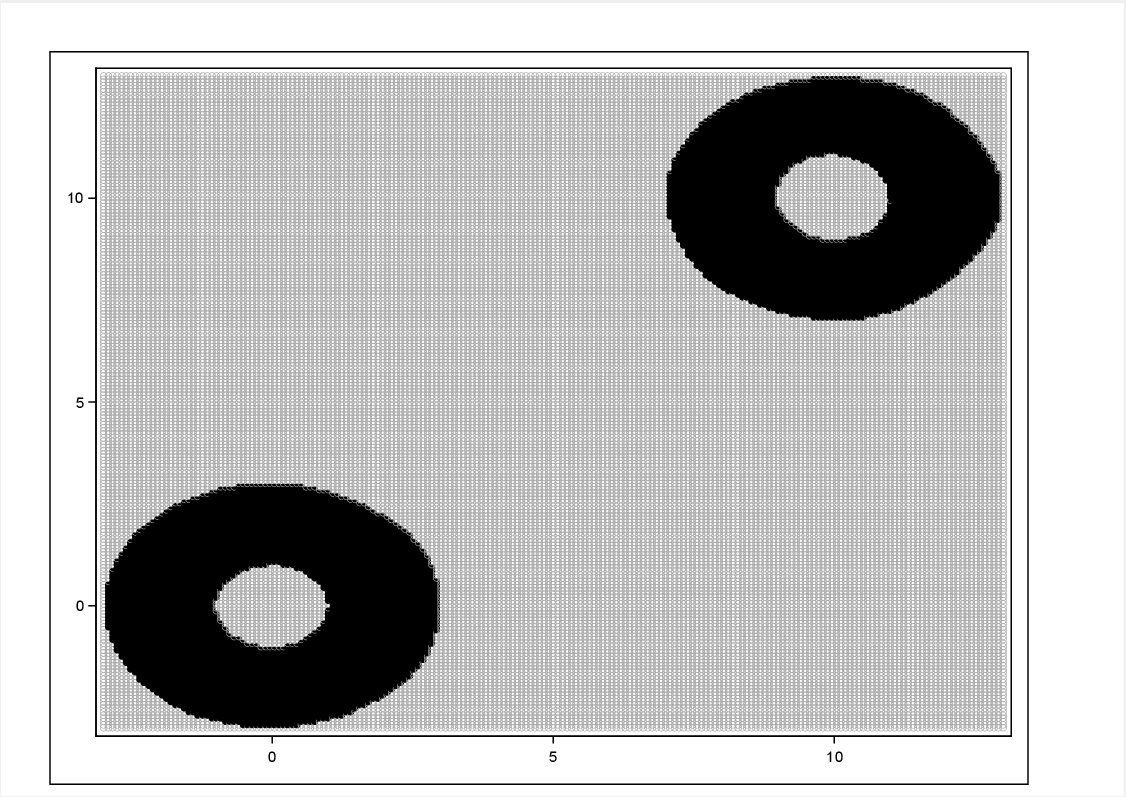} \\
(a)  & (b)  \\[6pt]
Full SVDD Method  & Sampling method \\[6pt]
\end{tabular}
\caption{Scoring results. Above figures show results of scoring on a 200x200 data grid. Light gray color indicates outside points and black color indicates inside points. Figure (a) used full SVDD method for training. Figure (b) used sampling method for training. }\label{fig:image_8}
\end{figure}

\begin{table}[h!]
\begin{minipage}{.5\textwidth}
    \begin{tabular}{||c c c c c||} 
    \hline
        Data & \#Obs & $R^{2}$ & \#SV & Time \\ [0.5ex] 
        \hline\hline
        Banana & 11,016  & 0.8789 & 21 & 1.98 sec   \\ 
        TwoDonut &1,333,334  & 0.8982 &178  & 32 min \\ 
        Star & 64,000 & 0.9362  &76  &11.55 sec   \\ [1ex] 
    \hline
    \end{tabular}
    \caption{SVDD Training using full SVDD method}\label{table:t2}
\end{minipage}\\
\mbox{}\\
\begin{minipage}{.4\textwidth}
	\begin{tabular}{||ccccc||} 
		\hline
		Data&Iterations & $R^{2}$ & \#SV & Time\\ [0.5ex] 
		\hline\hline
		Banana(6)&119&0.872&19&0.32 sec\\ 
		TwoDonut(11)&157&0.897&37&0.29 sec\\
		Star(11)&141&0.932&44&0.28 sec\\[1ex] 
		\hline
	\end{tabular}
	\caption{SVDD Results using Sampling Method (sample size in parenthesis)}\label{table:t3} 
\end{minipage}
\end{table}
Results provided in Table \ref{table:t2} and Table \ref{table:t3} indicate
that our method provides an order of magnitude performance improvement as
compared to training using all observations in a single iteration. The threshold
$R^{2}$ values obtained using the sampling-based method are approximately equal
to the values that can be obtained by training using all observations in a
single iteration. Although the radius values are same, to confirm if the data
boundary defined using support vectors is similar, we performed scoring on a
$200\times200$ data grid. Figure \ref{fig:image_8} provides the scoring results
for all data sets. The scoring results for the Banana-shaped and the Two-Donut-shaped are very similar
for both the method, the scoring results for the Star-shaped shaped data for the two methods are also similar
except for a region near the center.

\section{Analysis of High Dimensional Data}
\label{pd}
Section \ref{res} provided comparison of our sampling method with full SVDD
method. For two-dimensional data sets the performance of sampling method can
be visually judged using the scoring results. We tested the sampling method
with high dimensional datasets, where such visual feedback about classification
accuracy of sampling method is not available. We compared classification
accuracy of the sampling method with the accuracy of training with full SVDD
method. We use the $F_{1}$-measure to quantify the classification accuracy
\cite{zhuang2006parameter}.
The $F_{1}$-measure is defined as follows:
\begin{equation}  
F_{1}=\dfrac{2\times \text{Precision}\times \text{Recall}}{\text {Precision}+\text {Recall}},
\end{equation}
where:
\begin{align}
\text {Precision}=\dfrac{\text{true positives}}{\text{true positives} + \text{false positives}}\\
\text {Recall}=\dfrac{\text{true positives}}{\text{true positives} + \text{false negatives}}.
\end{align} 
Thus high precision relates to a low false positive rate, and high recall
relates to a low false negative rate. We chose the $F_{1}$-measure because it is
a composite measure that takes into account both the Precision and the Recall.
Models with higher values of $F_{1}$-measure provide a better fit. \\

\subsection{Analysis of Shuttle Data}
\begin{figure}[h]
	\centering
	\includegraphics[width=3.45in]{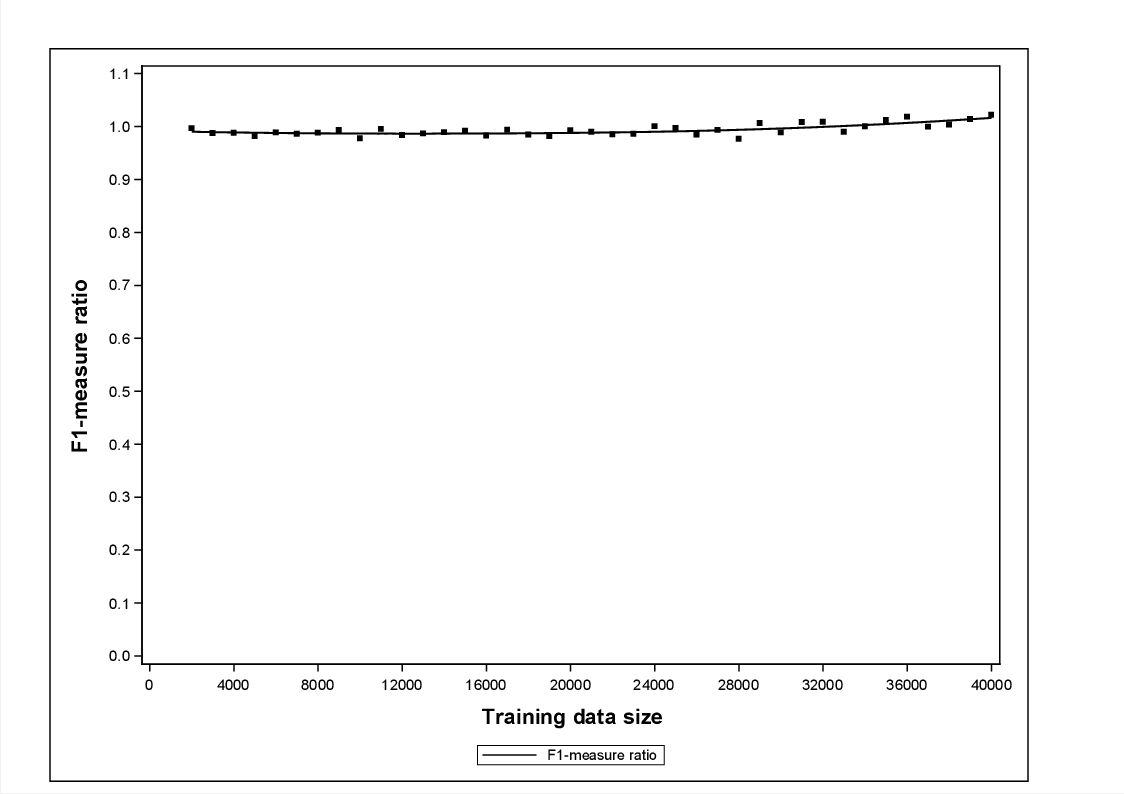}
	\caption{$F_1$-measure plot: Shuttle data. \textit{Sample size for sampling method=10}}
	\label{fig:image_9_1_1}
\end{figure}

\begin{figure}[h]
\centering
\includegraphics[width=3.45in]{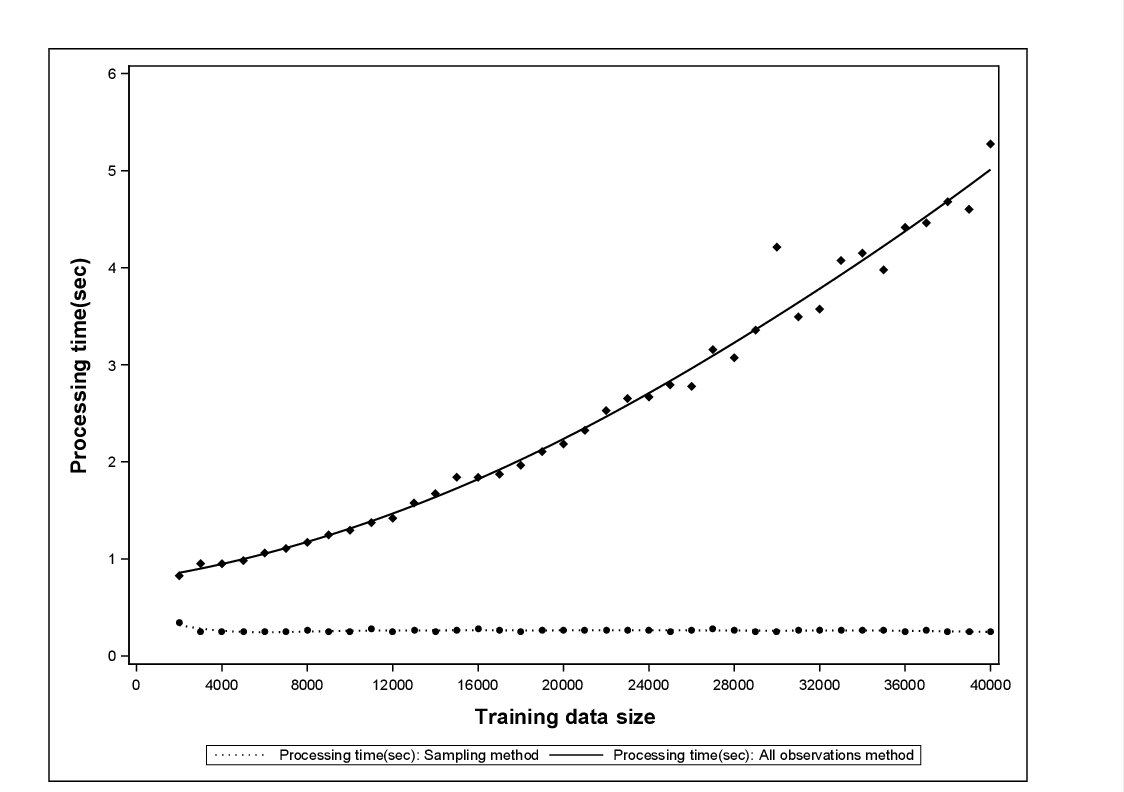}
\caption{Processing time plot: Shuttle data. \textit{Sample size for sampling method=10}}
\label{fig:image_9_1_2}
\end{figure}

In this section we provide results of our experiments with Statlog (shuttle)
dataset \cite{Lichman:2013}. This is a high dimensional data consists of nine
numeric attributes and one class attribute. Out of 58,000 total observations,
80\% of the observations belong to class one.
We created a training data set of randomly selected 2,000 observations belonging
to class one. The remaining 56,000 observations were used to create a scoring
data set. SVDD model was first trained using all observations in the training
data set. The training results were used to score the observations in the
scoring data set to determine if the model could accurately classify an
observation as belonging to class one and the accuracy of scoring was measured
using the $F_{1}$-measure. We then trained using the sampling-based method,
followed by scoring to compute the $F_{1}$-measure again. The sample size for
the sampling-based method was set to 10 (number of variables + 1). We measured
the performance of the sampling method using the $F_{1}$-measure ratio defined
as $F_{\text{Sampling}}/F_{\text{Allobs}}$ where
$F_{\text{Sampling}}$ is the $F_1$-measure obtained when the value obtained using the sampling method for training, and
$F_{\text{Allobs}}$ is the value of $F_1$-measure computed when all observations were used for training. A value close to 1 indicate that sampling method is competitive with full SVDD method.
We repeated the above steps varying the training data set of size from 3,000
to 40,000 in the increments of 1,000. The corresponding scoring data set size
changed from 55,000 to 18,000. Figure \ref{fig:image_9_1_1} provides the plot of
$F_{1}$-measure ratio. The plot of $F_1$-measure ratio is constant, very close
to 1 for all training data set sizes, provides the evidence that our sampling
method provides near identical classification accuracy as compared to full SVDD
method. Figure \ref{fig:image_9_1_2} provides the plot of the processing time
for the sampling method and training using all obsrvations. As the training
data set size increased, the processing time for full SVDD method increased
almost linearly to a value of about 5 seconds for training data set of 40,000
observations. In comparison, the processing time of the sampling based method
was in the range of 0.24 to 0.35 sec. The results prove that the sampling-based
method is efficient and it provides near identical results to full SVDD method.

\subsection{Analysis of Tennessee Eastman Data}
In this section we provide results of our experiments with high dimensional
Tennessee Eastman data. The data was generated using the MATLAB simulation code
\cite{Ricker:2002} which provides a model of an industrial chemical process
\cite{downs1993plant}. The data was generated for normal operations of the
process and twenty faulty processes. Each observation consists of 41 variables,
out of which 22 are measured continuously, on an average, every 6 seconds
and remaining 19 sampled at a specified interval either every 0.1 or 0.25
hours. We interpolated the 22 observations which are measured continuously
using SAS\textsuperscript{\textregistered} \textit{EXPAND} procedure. The
interpolation increased the observation frequency and generated 20 observations
per second. The interpolation ensured that we have adequate data volume to
compare performance our sampling method with full SVDD method.

We created a training data set of 5,000 randomly selected observations belonging
to the normal operations of the process. From the remaining observations, we
created a scoring data of 228,000 observations by randomly selecting 108,000
observations belonging to the normal operations and 120,000 observations
belonging to the faulty processes. A SVDD model was first trained using all
observations in the training data set. The training results were used to score
the observations in the scoring data set to determine if the model could
accurately classify an observation as belonging to the normal operations. The
accuracy of scoring was measured using the $F_{1}$-measure. We then trained
using the sampling method, followed by scoring to compute the $F_{1}$-measure
again. The sample size for the sampling based method was set to 42 (number
of variables + 1). Similar to the Shuttle data analysis, we measured the
performance of the sampling method using the $F_{1}$-measure ratio defined as
$F_{\text{Sampling}}/F_{\text{Allobs}}$ where $F_{\text{Sampling}}$ is the
$F_1$-measure obtained when the value obtained using the sampling method for
training, and $F_{\text{Allobs}}$ is the value of $F_1$-measure computed when
all observations were used for training. A value close to 1 indicate that
sampling method is competitive with full SVDD method.

\begin{figure}
	\centering
	\includegraphics[width=3.45in]{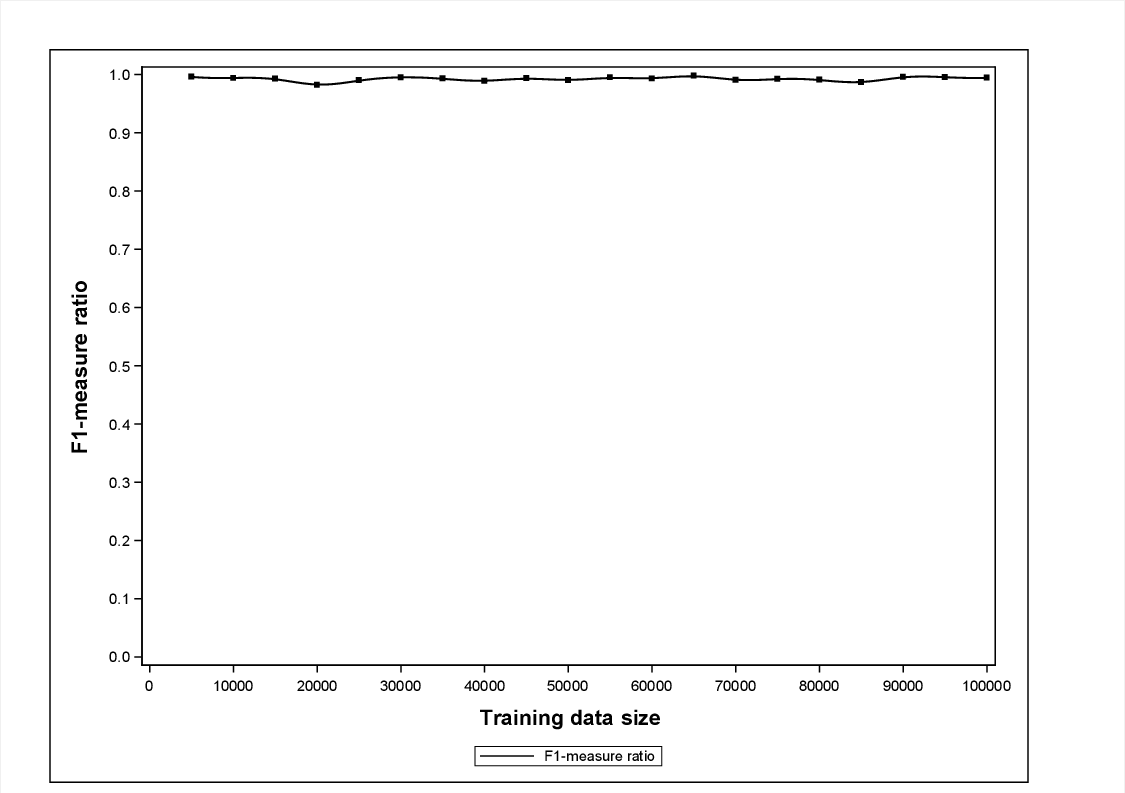}
	\caption{$F_1$-measure ratio plot: Tennessee Eastman data. Sample size for sampling method=42}
	\label{fig:image_9_1}
\end{figure}

\begin{figure}
\centering
\includegraphics[width=3.45in]{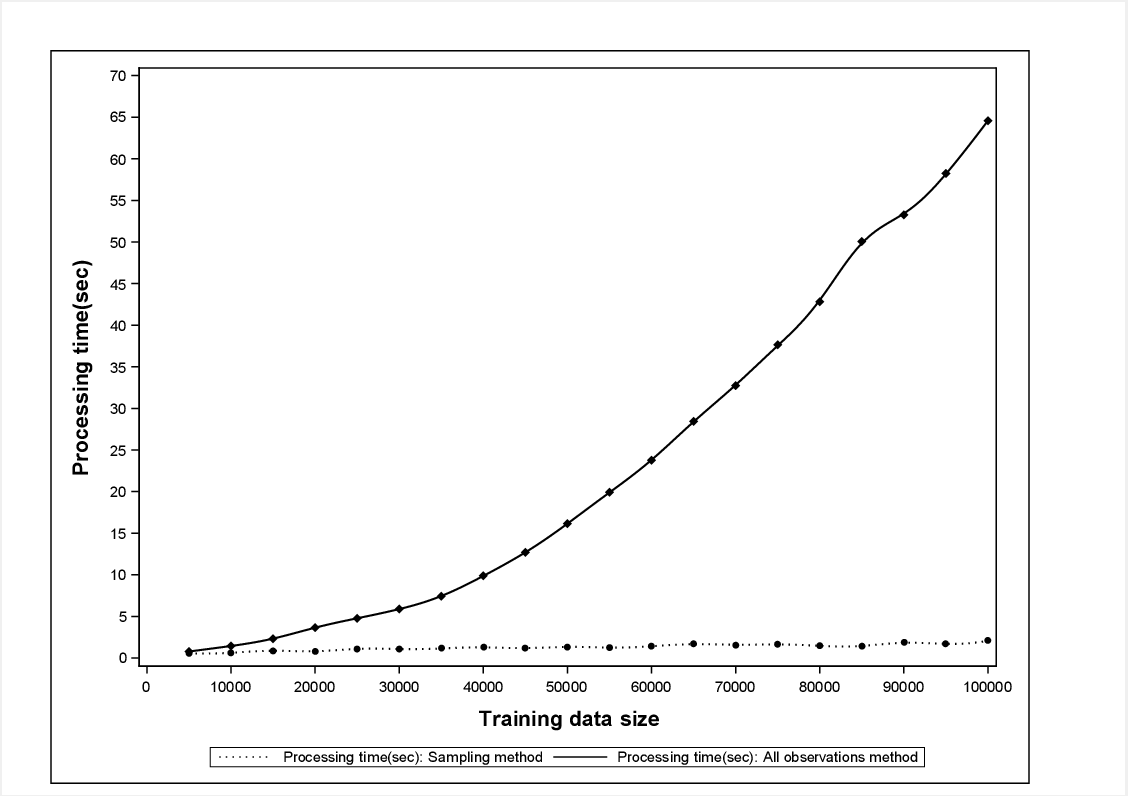}
\caption{Processing time plot: Tennessee Eastman data. Sample size for sampling method=42}
\label{fig:image_9_2}
\end{figure}

We repeated the above steps varying the training data set of size from 10,000
to 100,000 in the increments of 5,000. The scoring data set was kept unchanged
during each iteration. Figure \ref{fig:image_9_1} provides the plot of
$F_{1}$-measure ratio. The plot of $F_{1}$-measure ratio was constant, very
close to 1 for all training data set sizes, provides the evidence that the
sampling method provides near identical classification accuracy as compared
to full SVDD method. Figure \ref{fig:image_9_2} provides the plot of the
processing time for the sampling-based method and the all obsrvation method. As
the training data set size increased, the processing time for full SVDD method
increased almost linearly to a value of about one minute for training data set
of 100,000 observations. In comparison, the processing time of the sampling
based method was in the range of 0.5 to 2.0 sec. The results prove that the
sampling-based method is efficient and it provides and closely approximates
the results obtained from full SVDD method.

\section{Simulation Study}
\label{ss}
In this section we measure the accuracy of Sampling method when it is
applied to randomly generated polygons. Given the number of vertices, $k$,we
generate the vertices of a randomly generated polygon in the anticlockwise
sense as $r_1\exp i \theta_{(1)}, \dots, r_k \exp i \theta_{(k)}.$ Here
$\theta_{(i)}$'s are the order statistics of an i.i.d sample uniformly
drawn from $(0,2\pi)$ and $r_i$'s are uniformly drawn from an interval
$[\text{r}_{\text{min}},\text{r}_{\text{max}}].$ For this simulation we chose
$\text{r}_{\text{min}}=3$ and $\text{r}_{\text{max}}=5$ and varied the number
of vertices from $5$ to $30$. We generated $20$ random polygons for each vertex
size. Figure \ref{fig:image_10} shows two random polygons. Having determined a
polygon we randomly selected $600$ points uniformly from the interior of the
polygon to construct a training data set.

To create the scoring data set we the divided the bounding rectangle of each
polygon into a $200 \times 200$ grid. We labeled each point on this grid
as an ``inside'' or an ``outside'' point. We then fit SVDD on the training
data set and scored the corresponding scoring data set and calculated the
$F_1$-measure. The process of training and scoring was first performed using the
full SVDD method, followed by the sampling method. For sampling method we
used sample size of 5. We trained and scored each instance of a polygon 10 times
by changing the value of the Gaussian bandwidth parameter, $s$. We used $s$
values from the following set:\linebreak $s=[1, 1.44, 1.88, 2.33, 2.77, 3.22, 3.66,
4.11, 4.55, 5].$

As in previous examples we used the $F_!$ measure ratio to judge the accuracy of the sampling method.

\begin{figure}
	\centering     %%% not \center
	\subfloat[Number of Vertices = 5]{\label{fig:image_10}\includegraphics[width=3.45in]{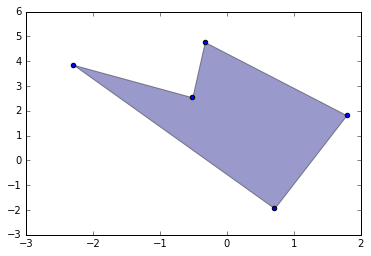}}\\
	\subfloat[Number of Vertices = 25]{\label{fig:image_11}\includegraphics[width=3.45in]{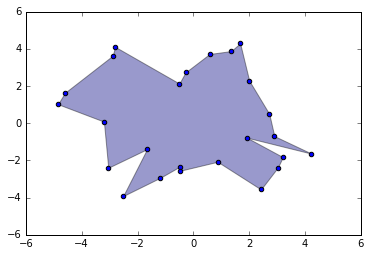}}
	\caption{Random Polygons}\label{fig:image_10}
\end{figure}

The Box-whisker plots in figures \ref{fig:image_11_a} to \ref{fig:image_11_c}
summarize the simulation study results. The x- axis shows the number of
vertices of the ploygon and y-axis shows the $F_1$-measure ratio. The bottom and
the top of the box shows the first and the third quartile values. The ends of
the whiskers represent the minimum and the maximum value of the $F_1$-measure
ratio. The diamond shape indicates the mean value and the horizontal line in the
box indicates the second quartile.
\subsection{Comparison of the best fit across $s$}
\begin{figure}
\centering
\includegraphics[width=3.45in]{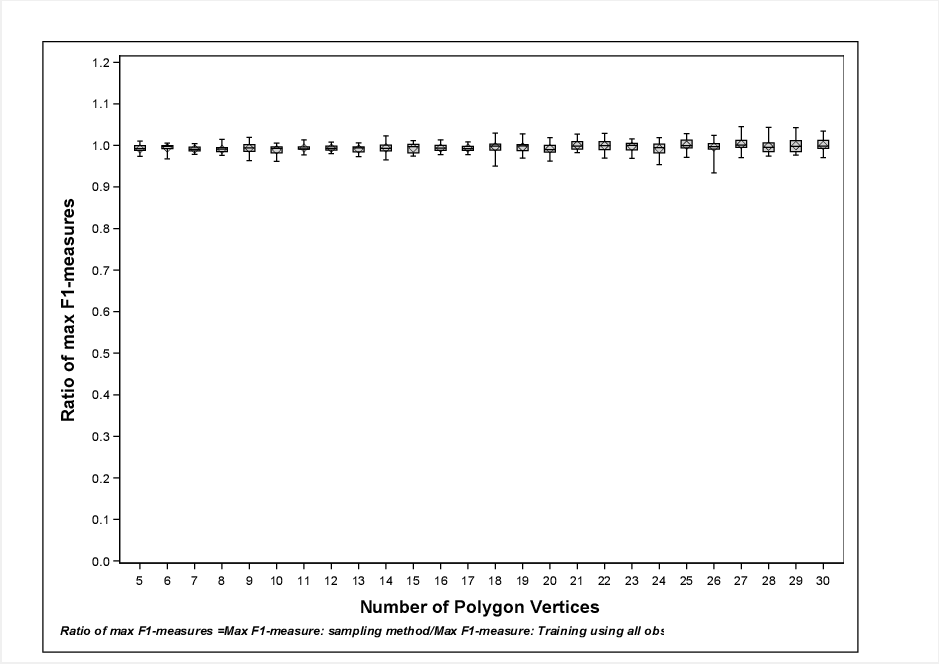}
\caption{Box-whisker plot: Number of vertices vs. Ratio of max $F_1$ measures \label{fig:image_11_a}}
\end{figure}

For each instance of a polygon we looked at $s$ value which provides the best
fit in terms of the $F_1$-ratio for each of the methods. The plot in Figure
\ref{fig:image_11_a} shows the plot of $F_{1}$ measure ratio computed using the
maximum values of $F_{1}$ measures. The plot shows that $F_1$-measure ratio
is greater than $\approx 0.92$ across all values of number of vertices. The
$F_1$ measure ratio in the top three quartiles is greater than $\approx$ 0.97
across all values of the number of vertices. Using best possible value of s, the
sampling method provides comparable results with full SVDD method.

\subsection{Results Using Same Value of $s$}
We evaluated sampling method against full SVDD method, for the same value
of $s$. The plots in Figure \ref{fig:image_11_b} illustrate the results for
different six different values of $s$. The plot shows that except for one outlier result in Figure \ref{fig:image_11_b} (d), $F_1$-measure
ratio is greater than 0.9 across number of vertices and $s$. In Figures
\ref{fig:image_11_b} (c) to (f), the top three quartiles of $F_{1}$ measure
ratio was consistently greater than $\approx 0.95$. Training using sampling method
and full SVDD method, using same $s$ value, provide similar results.\\

\begin{figure}
    \centering
	\subfloat[$s$=1]{\includegraphics[width=3.00in]{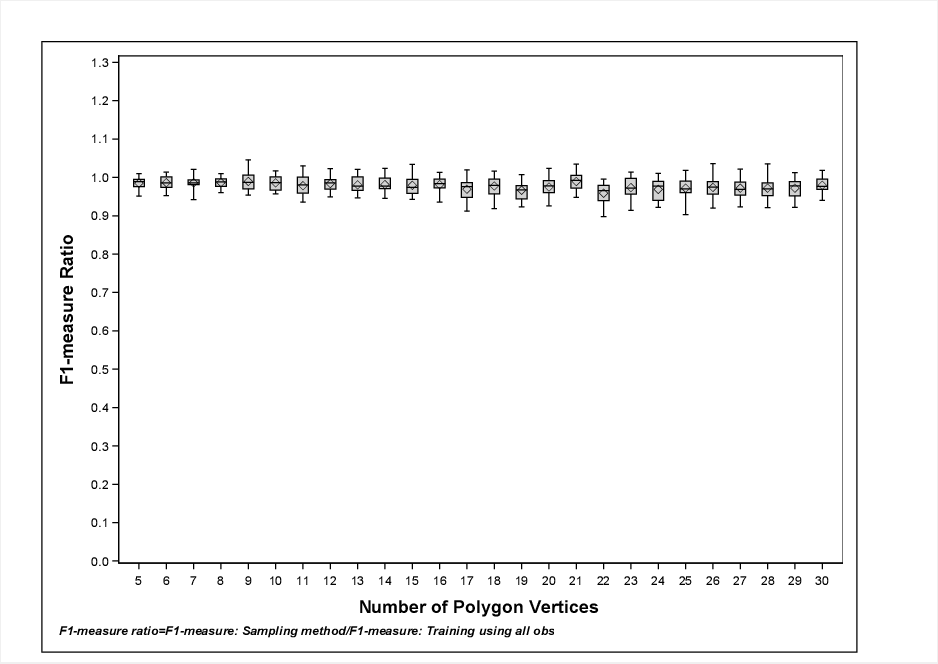}}\\
	\subfloat[$s$=1.4]{\includegraphics[width=3.00in]{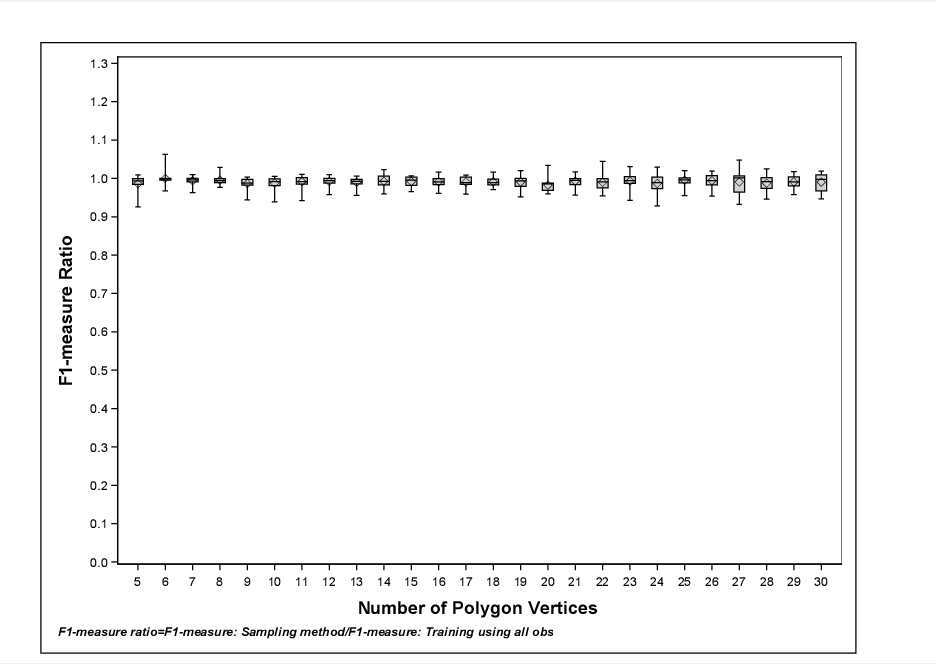}}\\
	\subfloat[$s$=2.3]{\includegraphics[width=3.00in]{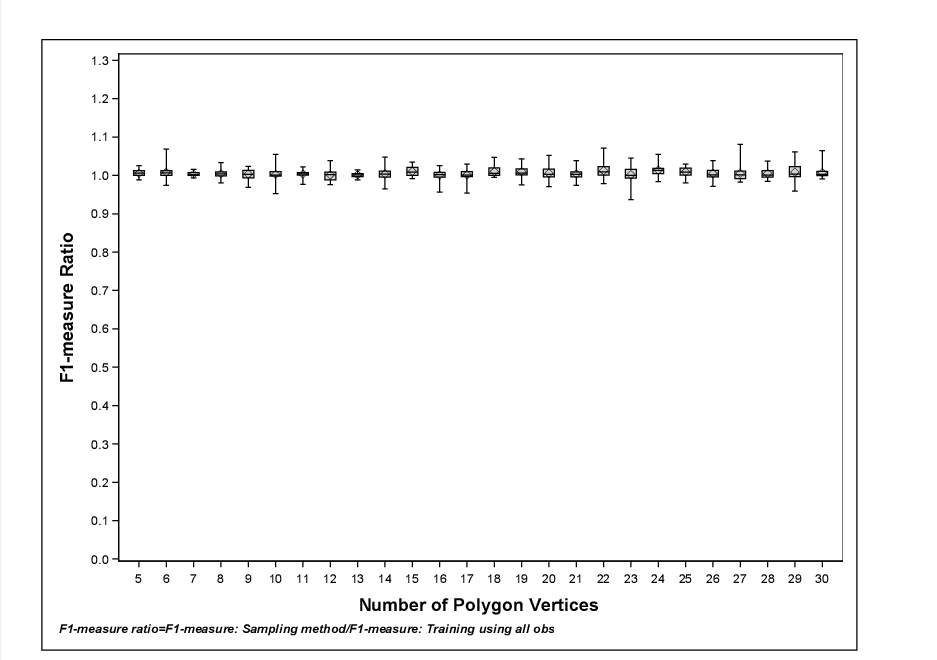}}\\
	\subfloat[$s$=3.4]{\includegraphics[width=3.00in]{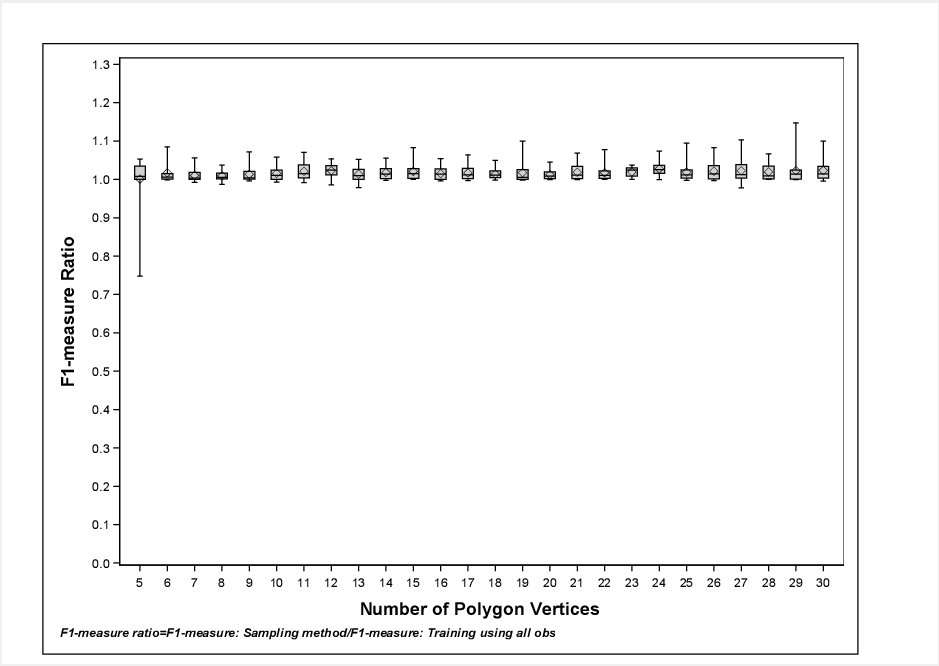}}
    \phantomcaption
\end{figure}
\begin{figure}
    \ContinuedFloat
    \centering
	\subfloat[$s$=4.1]{\includegraphics[width=3.00in]{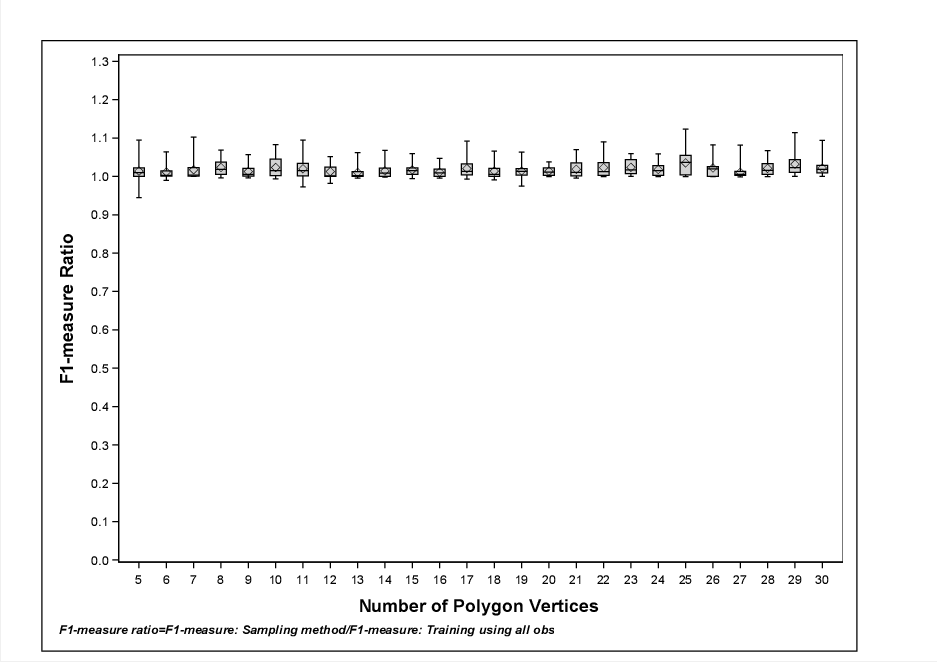}}\\
	\subfloat[$s$=5.0]{\includegraphics[width=3.00in]{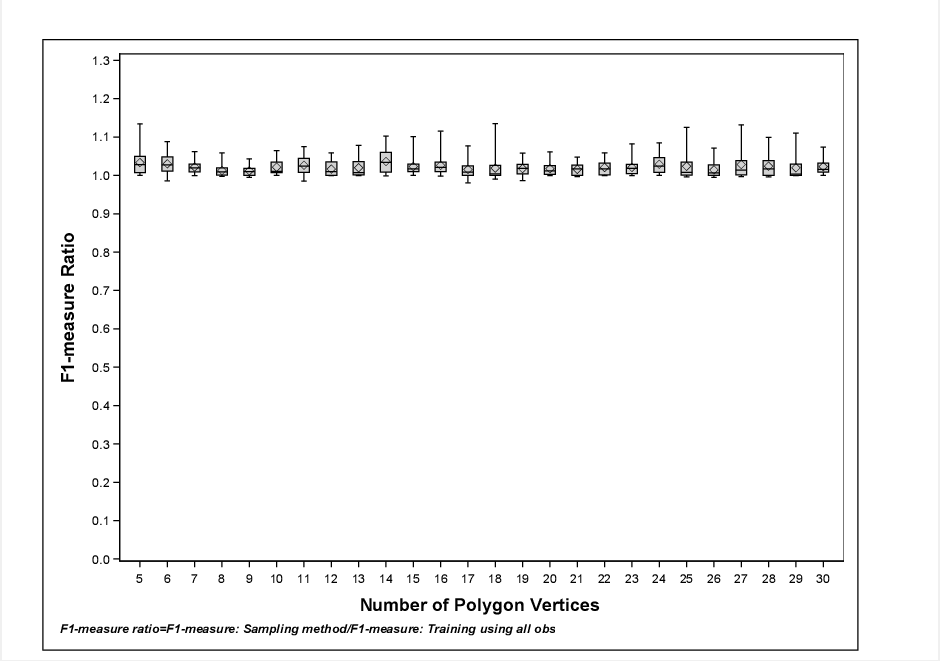}}
    \caption[]{Box-whisker plot: Number of vertices vs. $F_1$ measure ratio for different s values \label{fig:image_11_b}}
\end{figure}

\subsection{Overall Results}
\begin{figure}
\centering
\includegraphics[width=3.45in]{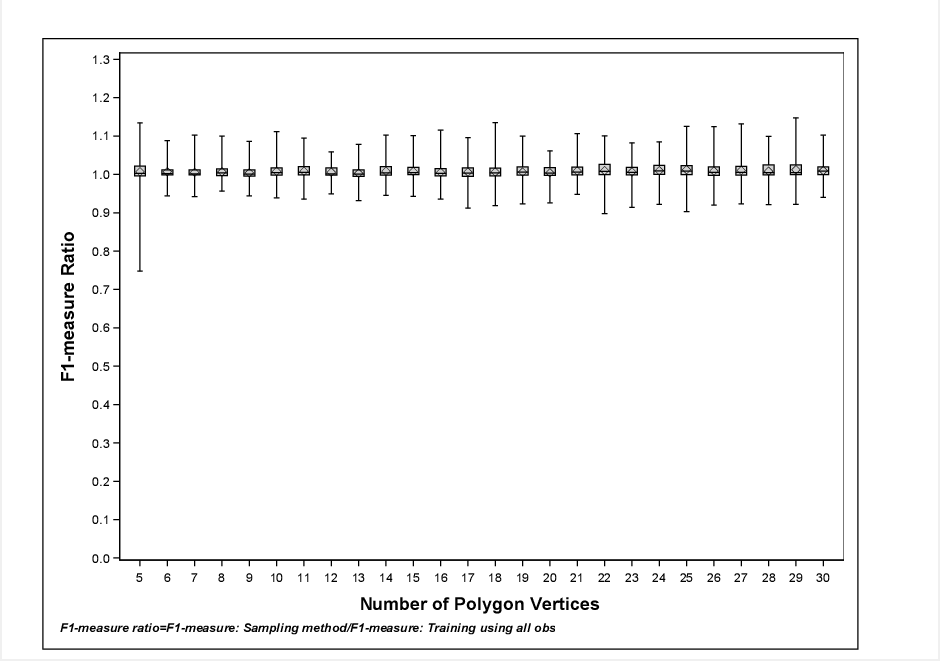}
\caption{Box-whisker plot: Number of vertices vs. $F_1$ measure ratio  }
\label{fig:image_11_c}
\end{figure}

Figure \ref{fig:image_11_c} provides summary of all simulation performed for
different polygon instances and varying values of $s$. The plot shows that
except for one outlier result, $F_1$-measure ratio is greater than 0.9 across number of vertice. The $F_1$
measure ratio in the top three quartiles is greater than $\approx 0.98$ across
all values of the number of vertices. The accuracy of sampling method is
comaprable to full SVDD method.

%\begin{table}[h!]
%\begin{minipage}{.5\textwidth}
%    \begin{tabular}{||c c c c c||} 
%    \hline
%        Data & \#Obs & $R^{2}$ & \#SV & Time \\ [0.5ex] 
%        \hline\hline
%        Banana & 11,016  & 0.8789 & 21 & 1.98 sec   \\ 
%        TwoDonut &1,333,334  & 0.8982 &178  & 32 min \\ 
%        Star & 64,000 & 0.9362  &76  &11.55 sec   \\ [1ex] 
%    \hline
%    \end{tabular}
%    \caption{SVDD Training using full SVDD method}\label{table:t2}
%\end{minipage}\\
%\mbox{}\\
%\begin{minipage}{.4\textwidth}
%	\begin{tabular}{||ccccc||} 
%		\hline
%		Data&Iterations & $R^{2}$ & \#SV & Time\\ [0.5ex] 
%		\hline\hline
%		Banana(6)&119&0.872&19&0.32 sec\\ 
%		TwoDonut(11)&157&0.897&37&0.29 sec\\
%		Star(11)&141&0.932&44&0.28 sec\\[1ex] 
%		\hline
%	\end{tabular}
%	\caption{SVDD Results using Sampling Method (sample size in parenthesis)}\label{table:t3} 
%\end{minipage}
%\end{table}

\section{Conclusion}
\label{cn}
We propose a simple sampling-based iterative method for training SVDD. The
method incrementally learns during each iteration by utilizing information
contained in the current master set of support vectors and new information
provided by the random sample. After a certain number of iterations, the
threshold $R^2$ value and the center $a$ start to converge. At this point, the
SVDD of the master set of support vectors is close to the SVDD of training data
set. We provide a mechanism to detect convergence and establish a stopping
criteria. The simplicity of proposed method ensures ease of implementation.
The implementation involves writing additional code for calling SVDD training
code iteratively, maintaining a master set of support vectors and implementing
convergence criteria based on threshold $R^2$ and center $a$. We do not
propose any changes to the core SVDD training algorithm as outlined in section
\ref{mfsvdd}. The method is fast. The number of observations used for
finding the SVDD in each iteration can be a very small fraction of the number
of observations in the training data set. The algorithm provides good results
in many cases with sample size as small as $m+1$, where $m$ is the number of variables in
the training data set. The small sample size ensures that each iteration of the
algorithm is extremely fast. The proposed method provides a fast alternative
to traditional SVDD training method which uses information from all observations
in one iteration. Even though the sampling based method provides an approximation 
of the data description but in applications where training data set is large, fast
approximation is often preferred to an exact description which takes more time to determine.
Within the broader realm of Internet of Things (IoT) we
expect to see multiple applications of SVDD especially to monitor industrial
processes and equipment health and many of these applications will require fast periodic training
using large data sets. This can be done very efficiently with our
method.
\bibliographystyle{plain}
\bibliography{svdd_sampling}

\begin{thebibliography}{10}

\bibitem{smsvddg}
Anonymous github account with a sample based svdd implementation in python.
\newblock \url{https://github.com/samplesvdd/sample_svdd}.

\bibitem{chang2011libsvm}
Chih-Chung Chang and Chih-Jen Lin.
\newblock Libsvm: a library for support vector machines.
\newblock {\em ACM Transactions on Intelligent Systems and Technology (TIST)},
  2(3):27, 2011.

\bibitem{downs1993plant}
James~J Downs and Ernest~F Vogel.
\newblock A plant-wide industrial process control problem.
\newblock {\em Computers \& chemical engineering}, 17(3):245--255, 1993.

\bibitem{ege2015IoT}
Gul Ege.
\newblock Multi-stage modeling delivers the roi for internet of things.
\newblock
  \url{http://blogs.sas.com/content/subconsciousmusings/2015/10/09/multi-stage-modeling-delivers-the-roi-for-internet-of-things/-…-is-epub/},
  2015.

\bibitem{kim2007fast}
Pyo Kim, Hyung Chang, Dong Song, and Jin Choi.
\newblock Fast support vector data description using k-means clustering.
\newblock {\em Advances in Neural Networks--ISNN 2007}, pages 506--514, 2007.

\bibitem{Lichman:2013}
M.~Lichman.
\newblock {UCI} machine learning repository, 2013.

\bibitem{luo2010fast}
Jian Luo, Bo~Li, Chang-qing Wu, and Yinghui Pan.
\newblock A fast svdd algorithm based on decomposition and combination for
  fault detection.
\newblock In {\em Control and Automation (ICCA), 2010 8th IEEE International
  Conference on}, pages 1924--1928. IEEE, 2010.

\bibitem{scikit-learn}
F.~Pedregosa, G.~Varoquaux, A.~Gramfort, V.~Michel, B.~Thirion, O.~Grisel,
  M.~Blondel, P.~Prettenhofer, R.~Weiss, V.~Dubourg, J.~Vanderplas, A.~Passos,
  D.~Cournapeau, M.~Brucher, M.~Perrot, and E.~Duchesnay.
\newblock Scikit-learn: Machine learning in {P}ython.
\newblock {\em Journal of Machine Learning Research}, 12:2825--2830, 2011.

\bibitem{Ricker:2002}
N.~Lawrence Ricker.
\newblock Tennessee eastman challenge archive, matlab 7.x code, 2002.
\newblock [Online; accessed 4-April-2016].

\bibitem{sanchez2007one}
Carolina Sanchez-Hernandez, Doreen~S Boyd, and Giles~M Foody.
\newblock One-class classification for mapping a specific land-cover class:
  Svdd classification of fenland.
\newblock {\em Geoscience and Remote Sensing, IEEE Transactions on},
  45(4):1061--1073, 2007.

\bibitem{sukchotrat2009one}
Thuntee Sukchotrat, Seoung~Bum Kim, and Fugee Tsung.
\newblock One-class classification-based control charts for multivariate
  process monitoring.
\newblock {\em IIE transactions}, 42(2):107--120, 2009.

\bibitem{tax2004support}
David~MJ Tax and Robert~PW Duin.
\newblock Support vector data description.
\newblock {\em Machine learning}, 54(1):45--66, 2004.

\bibitem{widodo2007support}
Achmad Widodo and Bo-Suk Yang.
\newblock Support vector machine in machine condition monitoring and fault
  diagnosis.
\newblock {\em Mechanical Systems and Signal Processing}, 21(6):2560--2574,
  2007.

\bibitem{ypma1999robust}
Alexander Ypma, David~MJ Tax, and Robert~PW Duin.
\newblock Robust machine fault detection with independent component analysis
  and support vector data description.
\newblock In {\em Neural Networks for Signal Processing IX, 1999. Proceedings
  of the 1999 IEEE Signal Processing Society Workshop.}, pages 67--76. IEEE,
  1999.

\bibitem{zhuang2006parameter}
Ling Zhuang and Honghua Dai.
\newblock Parameter optimization of kernel-based one-class classifier on
  imbalance learning.
\newblock {\em Journal of Computers}, 1(7):32--40, 2006.

\end{thebibliography}
\end{document}